%% file: main_arxiv.tex
\documentclass[11pt,a4paper]{article}
\PassOptionsToPackage{table}{xcolor}

\usepackage{hcl}
\setHCLLogo{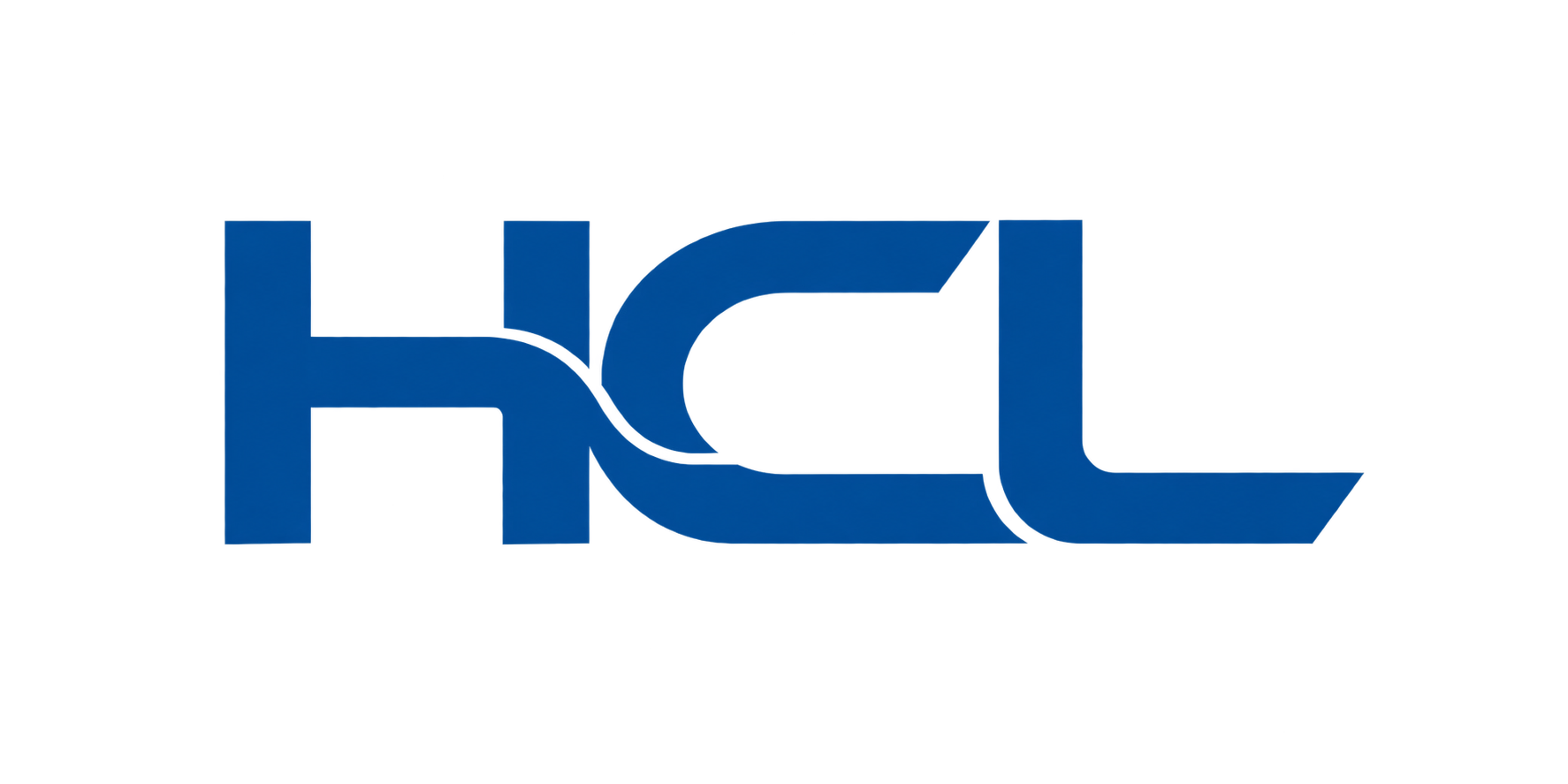}

\usepackage{amsmath}
\usepackage{pifont}      
\usepackage[authoryear,round,semicolon,sort]{natbib}
\usepackage{booktabs}
\usepackage{adjustbox}
\usepackage{float}
\usepackage{multirow}
\usepackage[normalem]{ulem}
\usepackage{wrapfig}
\usepackage{makecell}     
\usepackage{xcolor}       

\definecolor{lightpurple}{RGB}{230, 220, 235}
\definecolor{lightgreen}{RGB}{220, 230, 220}
\definecolor{lightpink}{RGB}{250, 240, 240}

\DeclareRobustCommand{\inlinecolorblock}[2]{%
  \begingroup
  \setlength{\fboxsep}{0.25pt}%
  \setlength{\fboxrule}{0.35pt}%
  \raisebox{-0.15ex}{\fcolorbox{#1}{#2}{\rule{0pt}{\fontcharht\font`X}\rule{0.45ex}{0pt}}}%
  \endgroup
}

\setEyebrow{Kimodo-Policy\,\textbullet\,Preprint 2026}
\usepackage[table]{xcolor}
\definecolor{lotuspink}{RGB}{240, 213, 218}

\setReportTag{Humanoid Computing Lab\,\textperiodcentered\,HKUST(GZ)}
\setHeaderTagline{}
\setUniversityLogo{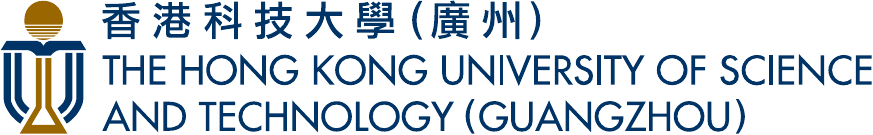}

\papertitle{\centering\fontsize{19pt}{23pt}\selectfont
  Kimodo-Policy: Learning Kinematic Action Models for\\
  Humanoid Loco-Manipulation}

\paperauthors{\centering%
  Yunheng Wang\equalmark\affilmark{1}\quad
  Jiaxi Zhang\equalmark\affilmark{1}\quad
  Wei Lu\equalmark\affilmark{1}\quad
  Yuetong Fang\affilmark{1}\quad
  Taowen Wang\affilmark{1}\quad
  Yixiao Feng\affilmark{1}\\[0.3em]
  Shuning Zhang\affilmark{1}\quad
  Junzhe Xu\affilmark{1,2}\quad
  Yichi Wang\affilmark{3}\quad
  Zizhao Yuan\affilmark{1}\quad
  Kailun Huang\affilmark{1}\\[0.3em]
  Lusong Li\affilmark{2}\quad
  Mingqian Zhou\affilmark{4}\quad
  Zecui Zeng\correspondmark\affilmark{2}\quad
  Renjing Xu\correspondmark\affilmark{1}%
}

\paperaffil{\centering%
  \affilmark{1}\,\textcolor{hclblue}{HKUST(GZ)}\quad
  \affilmark{2}\,\textcolor{hclblue}{JD Explore Academy}\quad
  \affilmark{3}\,\textcolor{hclblue}{BJUT}\quad
  \affilmark{4}\,\textcolor{hclblue}{Cambridge}
}

\papernotes{\centering%
  \equalmark\,Equal contribution (co-first authors).\quad
  \correspondmark\,Co-corresponding authors.%
}

\hypersetup{
  pdftitle={Kimodo-Policy: Learning Kinematic Action Models for Humanoid Loco-Manipulation},
  pdfauthor={Yunheng Wang, Jiaxi Zhang, Wei Lu, Yuetong Fang, Taowen Wang, Yixiao Feng, Shuning Zhang, Junzhe Xu, Yichi Wang, Zizhao Yuan, Lusong Li, Kailun Huang, Mingqian Zhou, Zecui Zeng, Renjing Xu}
}

\paperbadges{\centering\normalsize
  \href{https://kimodo-policy.github.io/}{\raisebox{-0.35em}{\includegraphics[height=1.3em]{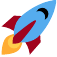}}\kern0.3em Project Page}\quad
  \href{https://github.com/kimodo-policy/kimodo-policy}{\raisebox{-0.35em}{\includegraphics[height=1.2em]{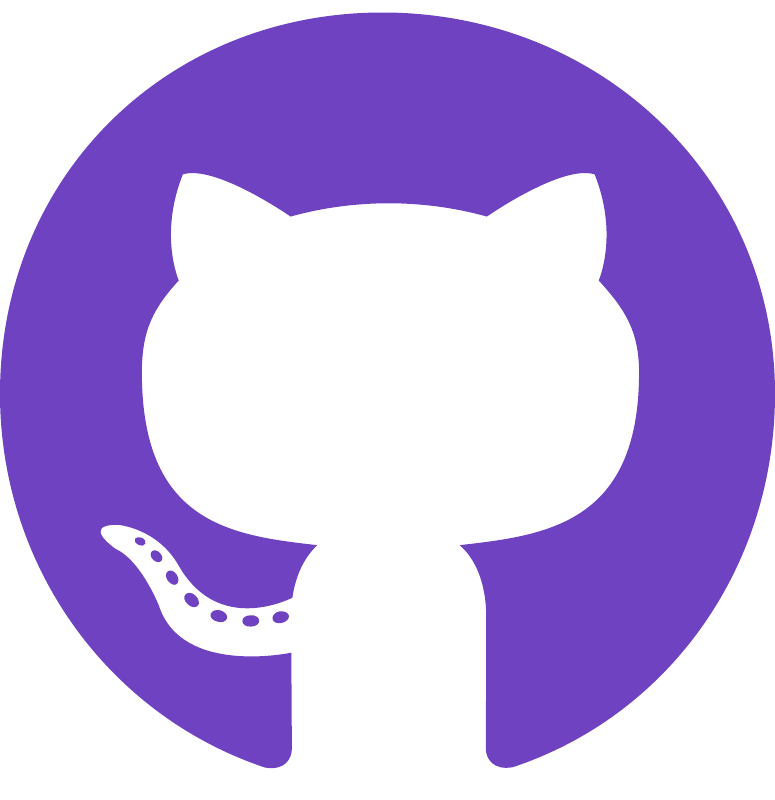}}\kern0.3em GitHub}\quad
  \href{https://huggingface.co/YunhengWang/kimodo-policy/tree/main}{\raisebox{-0.35em}{\includegraphics[height=1.2em]{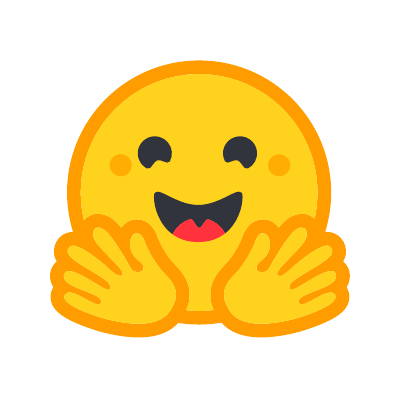}}\kern0.3em Hugging Face}\par
}

\begin{document}

\maketitleHCL

\input{content/Abstract}

\section{Introduction}
\label{sec:introduction}
\input{content/Introduction}

\input{content/Related_Work}

\section{Method: Kimodo-Policy}
\label{sec:method}
\input{content/Method}

\input{content/Experimental_Setup}

\input{content/Experiments_Results}

\input{content/Conclusion}


\bibliography{ref}

\end{document}

%% file: content/Abstract.tex
\begingroup
\setlength{\intextsep}{0pt}
\begin{figure}[H]
  \centering
  \captionsetup{skip=6pt}
  \includegraphics[width=\textwidth]{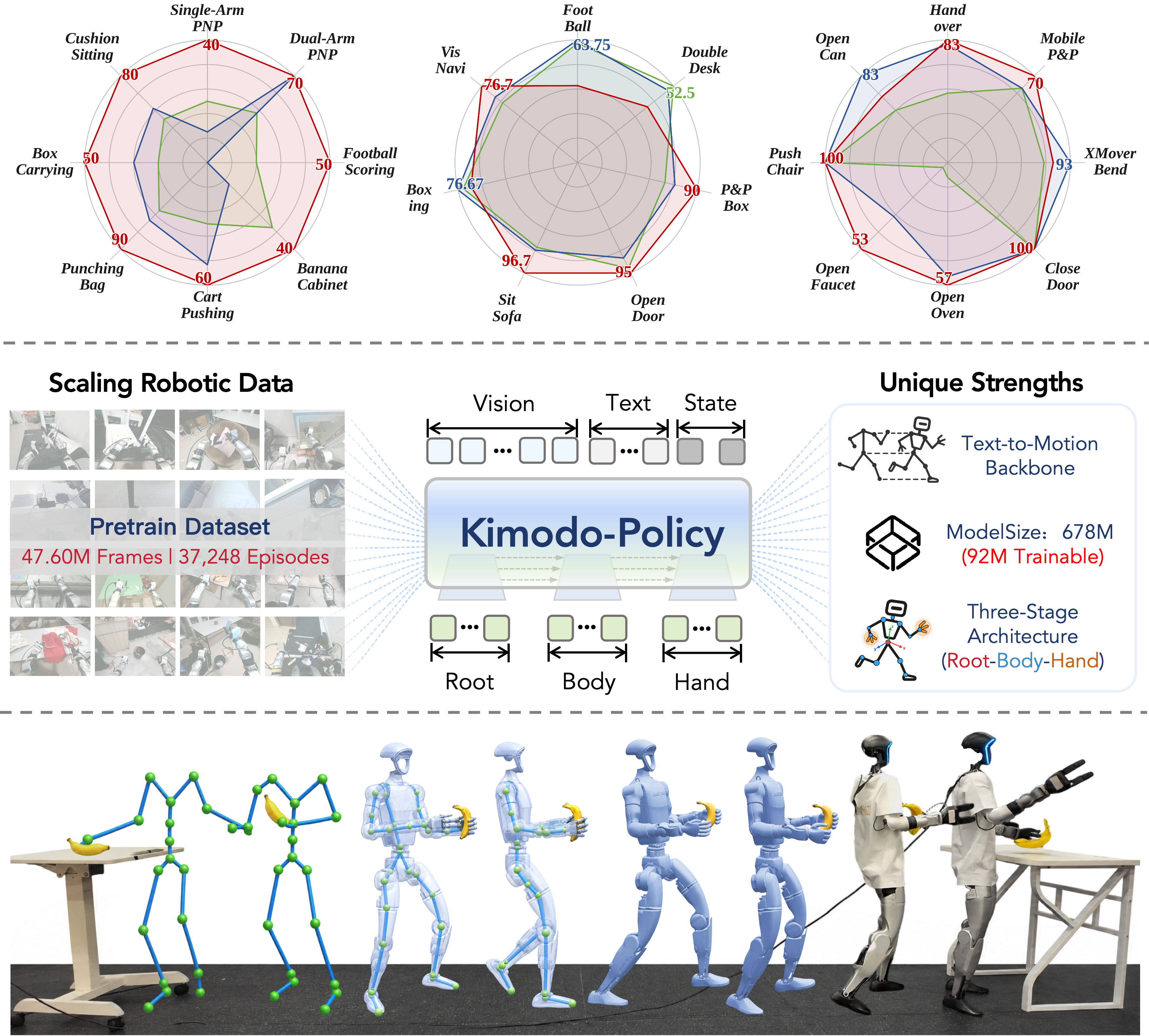}
  \caption{\textbf{Kimodo-Policy as an Instantiation of Kinematic Action Models (KAMs).}
  \textit{Top.} \inlinecolorblock{red!70!black}{red!70!black} Kimodo-Policy, \inlinecolorblock{blue!60!black}{blue!60!black} best baseline, and \inlinecolorblock{green!45!black!65!white}{green!45!black!65!white} second-best baseline. Our policy achieves higher mean success than the compared baselines on the displayed task sets.
  \textit{Middle.} Kimodo-Policy instantiates KAMs by inheriting whole-body motion priors from a pretrained motion generator. The compact policy (678M parameters, 92M trainable) combines large-scale loco-manipulation pretraining with a three-stage root--body--hand architecture.
  \textit{Bottom.} Observed motion history conditions future action-chunk generation for closed-loop control, as demonstrated in real-world object transfer.}
  \label{fig:overview}
\end{figure}
\endgroup

\begin{hclAbstract}
\noindent\textbf{Abstract.}\quad
Humanoid loco-manipulation requires continuous coordination of high-dimensional whole-body motion to realize task objectives. While vision--language--action models and world action models transfer semantic and visual-dynamics priors to robot control, these priors do not necessarily capture the structure of robot motion. We introduce \textbf{Kinematic Action Models (KAMs)}, a paradigm that treats whole-body motion as a direct pretraining target and uses a pretrained motion generator as the primary action-generation backbone. KAMs ground learned temporal and cross-body coordination priors through task conditioning, visual observations, and execution feedback. We present \textbf{Kimodo-Policy} as an initial instantiation, combining a frozen Kimodo text-to-motion backbone with lightweight ControlNet-based visual conditioning. A root--body--hand hierarchy coordinates locomotion and manipulation, while motion-history conditioning and repeated action-chunk prediction bridge motion synthesis and closed-loop control. Following large-scale loco-manipulation pretraining, the policy achieves the highest overall success among compared policies on multi-task HumanoidArena (72.14\%), competitive performance on SIMPLE, and 60\% mean success across eight real-world tasks, with 678M parameters (92M trainable). Ablations support the contribution of the pretrained motion backbone. Together, these results provide initial evidence for motion generation as a complementary foundation for humanoid action learning.
\end{hclAbstract}

%% file: content/Introduction.tex


\begingroup
\linespread{0.97}\selectfont
\setlength{\parskip}{0.25em plus 0.05em minus 0.03em}
\setlength{\textfloatsep}{10pt plus 2pt minus 2pt}
\suppressfloats[t]
\begin{figure*}[t]
  \centering
  \includegraphics[width=0.98\textwidth]{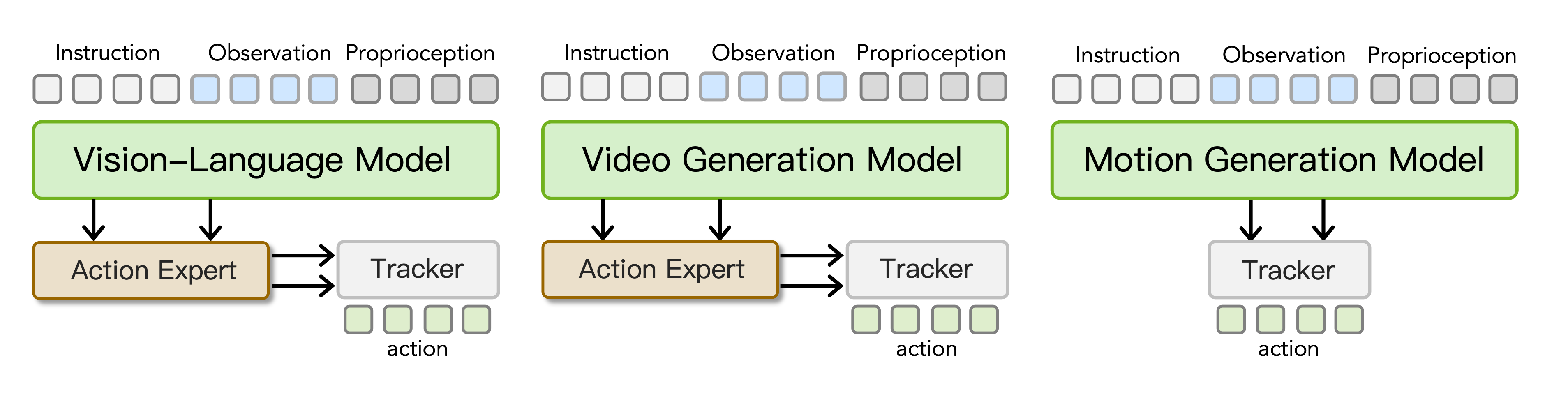}
  \caption{\textbf{Pretrained Backbones for Humanoid Action Generation.}
  From left to right: VLA, WAM, and KAM paradigms built around vision--language, video generation, and motion generation models.
  The schematic highlights their primary sources of generative priors: KAMs adapt a pretrained motion generator to produce kinematic action sequences for a downstream tracker, directly transferring learned movement structure into the policy.}
  \label{fig:policy_routes}
\end{figure*}

Humanoid loco-manipulation integrates task understanding, visual perception, and high-dimensional whole-body coordination~\citep{psi0,openhlm}. Beyond the requirements of conventional manipulation, it demands that robots continuously coordinate their root, torso, arms, and legs throughout an interaction~\citep{leverb}. For example, reaching for an object while moving requires a robot to simultaneously adjust its body position, maintain dynamic balance, and coordinate joint movements according to the object's location and contact state. Whole-body motion is therefore an integral part of accomplishing the task, rather than merely an execution stage that follows task-level decision-making. Even with accurate task understanding and perception, a robot cannot complete the task without generating temporally and spatially coordinated, executable whole-body motion. The central challenge of loco-manipulation thus extends beyond determining what the robot should do to generating the high-dimensional, continuous, and coupled whole-body motion required for task execution.

Recent advances in vision--language--action models (VLAs) and world action models (WAMs) offer two representative approaches to robot action generation. VLAs typically extend the semantic understanding and visual representations acquired by pretrained vision--language models to robot action prediction~\citep{rt2,openvla,pi0,grootn1}. WAMs draw on the visual dynamics learned by pretrained video generation models to connect future-state prediction with action generation~\citep{dreamzero,motionwam,wholebodywam_wbc}. These approaches provide important foundations for task understanding, environmental perception, and interaction planning. However, pretrained knowledge of visual semantics and dynamics does not necessarily capture the structural regularities of high-dimensional robot motion. Although these models can acquire whole-body motion capabilities through action supervision and downstream policy learning~\citep{wholebodyvla}, translating visual or semantic representations into coordinated whole-body motion still requires action representations and dependencies that reflect the structure of the robot's motion space. This motivates a complementary question: beyond semantic understanding and visual dynamics, can the structure of whole-body motion itself serve as a distinct target for generative pretraining?

Motivated by this observation, \textit{we introduce Kinematic Action Models (KAMs), a paradigm that uses a pretrained motion generation model as the primary backbone for action generation} (Fig.~\ref{fig:policy_routes}). The central premise of KAMs is that, for humanoid robots whose behavior is expressed through high-dimensional whole-body motion, motion should be a direct target of foundation-model pretraining, in addition to being the final output of a policy. Whereas visual and language pretraining primarily captures visual semantics or the evolution of visual states, motion generation pretraining directly models the temporal evolution of body configurations, learning patterns of joint motion, root motion, and coordination across body parts~\citep{mdm,kimodo}. This allows the model to acquire structural priors in a motion space closely aligned with the target action-generation problem, reducing its reliance on downstream task learning to acquire these basic motion \mbox{regularities}. KAMs then use the pretrained motion generator as the policy's generative backbone, conditioning it on task information, visual observations, and robot states to adapt its existing whole-body motion capabilities to specific environments and objectives. Motion generation thereby becomes a foundation for policy pretraining, rather than solely an output stage following task understanding.

Transferring pretrained motion generation capabilities to robot control nevertheless requires addressing the gap between motion synthesis and closed-loop interaction. Conventional motion generation models typically produce coherent motion sequences from text or kinematic constraints~\citep{mdm,omnicontrol,kimodo}, whereas loco-manipulation requires robots to continually revise their subsequent motion based on current observations and their own states. The challenge is therefore both to acquire priors that capture whole-body coordination and to preserve these priors while making action generation responsive to task requirements and environmental changes. Within this framework, the pretrained motion generator provides reusable whole-body motion structure, perceptual conditioning grounds motion predictions in the current scene and task, and motion history together with continually updated action predictions allows the generation process to incorporate execution feedback. By combining motion structure with task conditioning and closed-loop feedback, KAMs provide a path from general-purpose motion generation to embodied interaction and control.

To investigate this paradigm, we present \textbf{Kimodo-Policy}, an initial instantiation of KAMs for humanoid loco-manipulation (Fig.~\ref{fig:overview}). We combine Kimodo's pretrained text-to-motion backbone~\citep{kimodo} with a lightweight ControlNet-based visual adapter~\citep{zhang2023adding} to directly examine whether strong motion priors can support visually conditioned action generation. The motion backbone remains frozen, while the adapter injects egocentric visual features to guide its predictions. A three-stage root--body--hand architecture first generates a root trajectory, then produces body joint motion conditioned on that trajectory, and finally predicts hand control signals from body features and visual context. Large-scale loco-manipulation pretraining adapts this architecture to robot interaction, while motion-history conditioning and repeated action-chunk prediction support closed-loop execution through a whole-body tracking controller~\citep{sonic}. The resulting policy contains 678 million parameters, of which 92 million are trainable. Across simulation benchmarks and real-robot tasks, Kimodo-Policy achieves the highest overall success rate among the compared policies on multitask HumanoidArena~\citep{humanoidarena}, remains competitive on SIMPLE~\citep{simple}, and outperforms the evaluated baselines on real-world loco-manipulation tasks. Ablation studies support the contribution of the pretrained motion backbone, while comparisons across motion priors show that the choice of pretrained motion model affects downstream performance. These results provide initial evidence that motion generation can serve as a practical foundation for humanoid action models. Our main contributions are threefold:

\begin{samepage}
\begin{itemize}[leftmargin=*,topsep=0pt,itemsep=0.15em,parsep=0pt,partopsep=0pt]
  \item We introduce Kinematic Action Models (KAMs), a paradigm built on pretrained motion generators. By making whole-body motion a direct pretraining target, KAMs transfer temporal and cross-body coordination priors to downstream control, establishing a complementary foundation for humanoid policy learning.
  \item We develop Kimodo-Policy, an initial KAM instantiation with a frozen motion backbone and lightweight visual conditioning. Its root--body--hand hierarchy and motion-history conditioning support closed-loop loco-manipulation in a compact policy with 678M parameters, of which only 92M are trainable.
  \item We achieve the highest overall success among the compared methods on multitask HumanoidArena (72.14\%), competitive SIMPLE performance, and 60\% mean success across eight real-world tasks, outperforming the evaluated real-world baselines. Ablations substantiate the contribution of pretrained motion priors to humanoid control.
\end{itemize}
\end{samepage}
\endgroup

%% file: content/Related_Work.tex
\begingroup
\linespread{0.97}\selectfont
\section{Related Work}
\label{sec:related_work}

\subsection{Whole-Body Control for Loco-Manipulation}
\label{sec:related_whole_body}
Humanoid loco-manipulation integrates spatial grounding~\citep{wang2026limitsvisionandlanguagenavigation,uniground} and navigation planning~\citep{wang2025dreamnavtrajectorybasedimaginativeframework,hero} with whole-body coordination~\citep{leverb,openhlm}, enabling robots to approach interaction targets and manipulate objects while maintaining balance. HumanVLA~\citep{humanvla} distills a reinforcement-learned teacher into a vision--language policy. OASIS~\citep{oasis} combines simulated teleoperation with visual randomization, while FetchMan~\citep{fetchman} refines policies learned from synthetic demonstrations through reinforcement learning. Foundation policies such as GR00T N1~\citep{grootn1} and $\Psi_0$~\citep{psi0} transfer pretrained visual--semantic representations to robot action experts. LeVERB~\citep{leverb} and WholeBodyVLA~\citep{wholebodyvla} further investigate learned latent interfaces and loco-manipulation-oriented controllers that connect task understanding to whole-body execution. These developments highlight the complementary roles of perceptual generalization, interaction data, and the control interface. World action models (WAMs) additionally use future visual or latent-state prediction to inform action learning. MotionWAM~\citep{motionwam} conditions whole-body action generation on intermediate video-denoising features, whereas the WBC-grounded WholeBodyWAM of \citet{wholebodywam_wbc} adapts pretrained world--action priors through a unified controller interface and coordination-aware attention. More recent formulations address distinct aspects of predictive control: $\omega$-0~\citep{omega0} couples latent visual prediction with whole-body action generation, HWAM~\citep{hwam} jointly predicts reference actions and their realized body states, and Language-Forcing~\citep{li2026languageforcing} aligns predictive visual features with task-relevant language supervision. These approaches strengthen semantic grounding, predictive modeling, and execution awareness. However, pretraining on visual semantics or dynamics does not by itself provide a generative model of the coupled kinematics required for whole-body control. KAMs address this gap by making whole-body motion a direct target of generative pretraining and reusing the resulting generator as the action backbone. Perception then conditions an existing model of temporal and cross-body coordination, making coordinated movement a transferable foundation for humanoid policy learning.

\subsection{Motion Priors for Loco-Manipulation}
\label{sec:related_motion_priors}
Human motion provides supervision for coordination beyond what is available in costly robot demonstrations. Humanoid-VLA~\citep{humanoidvla} first aligns language and motion within an autoregressive language model, then incorporates egocentric visual context; ZeroWBC~\citep{zerowbc} adapts a VLM with image--text--motion data and executes the generated human motion through retargeting and interactive tracking. Predictive models offer another route. Being-M0.7~\citep{beingm07} learns a joint visual--motion prior from paired and unpaired human data, adapts it to robot trajectories, and conditions a separate action expert on its predictive representations. The motion-prior-based WholeBodyWAM of \citet{wholebodywam_motionpriors} first pretrains a motion expert on heterogeneous whole-body trajectories, then integrates video, motion, and action experts through joint post-training. These methods demonstrate the value of learning movement structure before target-task adaptation and of connecting that structure to visual context. The key distinction is how the learned motion prior enters action generation. Language-model-based approaches learn motion within a semantic backbone, while predictive-expert architectures align motion representations with a separate robot-action stream. KAMs reuse the pretrained motion generator itself as the primary action backbone: the model pretrained to generate coordinated movement also produces the kinematic sequences supplied to control. Visual observations, task information, and execution history condition this generative process, which accommodates different motion generators and perceptual encoders, including VLMs. This direct reuse allows robot-interaction learning to focus on grounding pretrained movement capabilities in the scene and task, with temporal and cross-body coordination already encoded in the backbone.

\subsection{Motion Generation}
\label{sec:related_motion_generation}
Motion generation extends beyond text-conditioned synthesis to action-conditioned animation, temporal completion, spatial control, and interaction with scenes and objects. ACTOR~\citep{actor} learns an action-conditioned variational motion representation; MDM~\citep{mdm} and MLD~\citep{mld} model motion with diffusion in pose and latent spaces, respectively. MoMask~\citep{momask} generates discrete motion tokens through masked modeling, while DartControl~\citep{dartcontrol} autoregressively composes diffusion-based motion primitives for online control. Controllability is developed along complementary axes: OmniControl~\citep{omnicontrol} constrains arbitrary joints over time, CHOIS~\citep{chois} jointly synthesizes human and object motion under language and waypoint conditions, and MaskedMimic~\citep{maskedmimic} unifies physics-based character control through motion inpainting with partial motion, text, and scene inputs. Together, these approaches show that motion models can encode reusable movement structure while responding to diverse conditioning signals. Kimodo~\citep{kimodo} advances this direction with large-scale motion-capture pretraining, a root--body diffusion hierarchy, and flexible constraints including keyframes, joint targets, and paths. Its released human and humanoid variants make controllable kinematic generation directly relevant to robot action modeling. However, controllable motion synthesis alone does not specify how a robot should revise its actions as the observed scene and executed motion change. Kimodo-Policy addresses this gap by coupling Kimodo's frozen generator with visual adaptation, coordinated hand prediction, and motion-history-conditioned replanning. Each policy query grounds future action chunks in current observations and anchors them to the robot's executed motion. Motion generation thereby becomes the mechanism for continually producing and revising robot actions, establishing a concrete KAM instantiation that preserves the pretrained backbone while learning its adaptation to interaction.
\par
\endgroup

%% file: content/Method.tex
This section presents Kimodo-Policy, an instantiation of KAMs for closed-loop humanoid loco-manipulation. We first introduce the pretrained motion backbone and visual conditioning mechanism (Sec.~\ref{sec:method_preliminaries}), then describe the root--body--hand architecture (Sec.~\ref{sec:method_architecture}). We next explain how motion-history conditioning adapts motion generation to closed-loop interaction (Sec.~\ref{sec:t2m_transfer}), and conclude with the training objective and inference procedure (Sec.~\ref{sec:training_inference}).

\subsection{Preliminaries}
\label{sec:method_preliminaries}

\textbf{Kimodo as a KAM Backbone.} Kimodo-Policy instantiates Kinematic Action Models (KAMs) by using a pretrained motion generator as the backbone for humanoid action generation. We adopt Kimodo-G1~\citep{kimodo}, which generates robot kinematic motion from language instructions and optional motion constraints through a hierarchical root--body denoising process. The root stage models global translation and heading, while the body stage generates articulated motion conditioned on the root prediction. Its motion representation jointly describes root motion, joint positions, velocities, orientations, and foot contacts, providing a structured space for whole-body prediction. Pretraining in this space allows the backbone to acquire temporal and cross-body coordination patterns that can be reused during downstream policy learning. We retain this pretrained generation structure and keep the backbone frozen, learning additional conditioning modules to adapt its predictions to robot interaction. Motion generation thus provides the foundation of the policy, while adapting it to closed-loop control requires grounding its predictions in the current scene and execution state.

\textbf{ControlNet for Visual Conditioning.} To make the pretrained motion generator responsive to the current scene, we adopt ControlNet-based conditioning~\citep{zhang2023adding} for the frozen root and body denoisers. Trainable control branches process egocentric visual features and inject residual signals into selected layers of the corresponding denoisers. The injection projections are initialized to zero, so the root--body generator initially follows its pretrained computation before learning to incorporate visual information. This design separates the pretrained motion backbone from the parameters learned for visual conditioning, allowing robot demonstrations to shape scene-dependent motion predictions without updating the backbone weights. Within this KAM instantiation, the motion generator supplies learned whole-body structure, while ControlNet provides a trainable interface through which observations guide action generation.

\subsection{Model Architecture}
\label{sec:method_architecture}

\textbf{Problem Formulation.} We formulate language-conditioned humanoid loco-manipulation as a closed-loop prediction problem in which the policy generates short, structured action chunks that combine root motion, articulated body motion, and hand controls. At environment step $k$, the policy receives a natural-language instruction $\ell$, the current egocentric RGB image $I_k$, an available motion history $\mathcal{H}_k=[\mathbf{x}_{k-L+1},\ldots,\mathbf{x}_k]$ encoded in the Kimodo representation, and the current bilateral hand state $\mathbf{s}^{h}_k$. It predicts a future chunk of $K$ frames, with each action $\mathbf{a}_i=[\mathbf{r}_i,\mathbf{b}_i,\mathbf{h}_i]$ comprising root motion, articulated body motion, and left/right hand-control signals:
\begin{equation}
  \mathbf{A}_{k+1:k+K}\sim\pi_{\theta}(\,\cdot\mid I_k,\ell,\mathcal{H}_k,\mathbf{s}^{h}_k).
  \label{eq:policy_formulation}
\end{equation}
The root and body components provide kinematic references for a low-level motion tracker~\citep{sonic}, while the hand component specifies synchronized closure commands with binary or continuous parameterizations according to the task interface. The policy executes a chunk prefix and replans from updated observations, using visual feedback and motion history to guide subsequent predictions.

\begin{figure}[H]
  \centering
  \includegraphics[width=\textwidth]{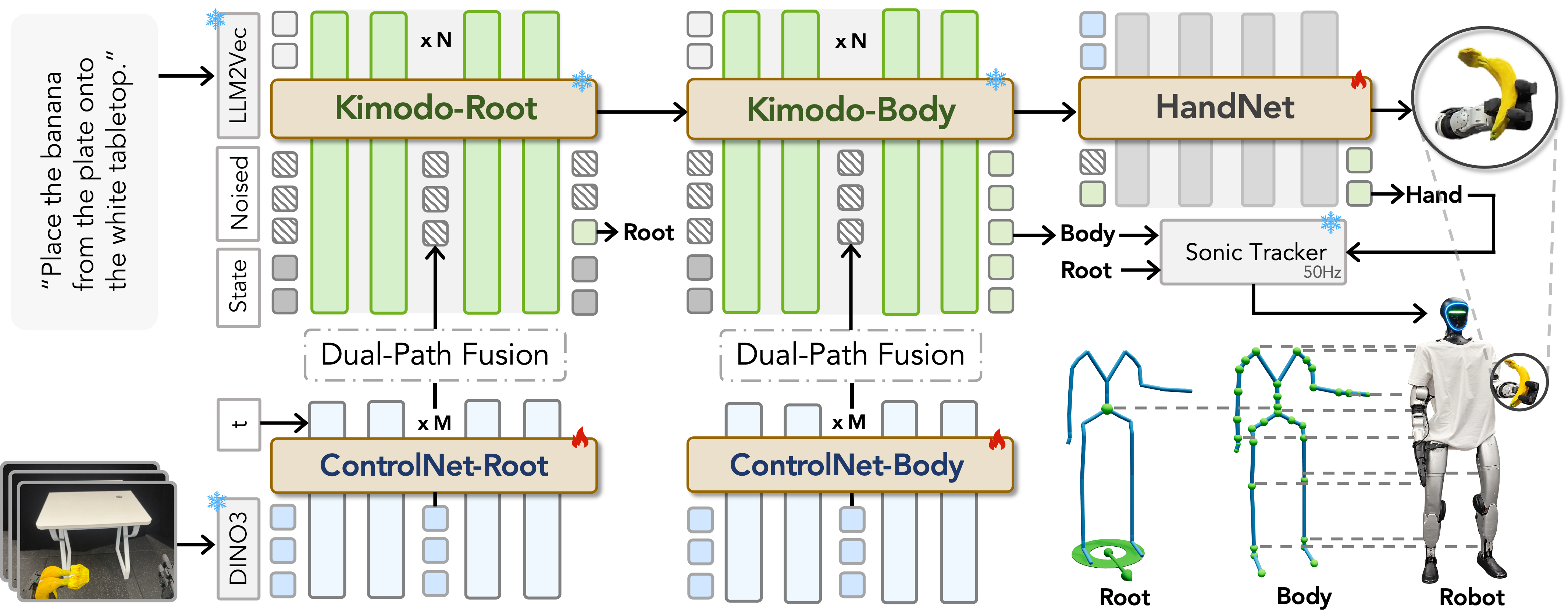}
  \caption{\textbf{Overview of Kimodo-Policy.} The colored vertical blocks denote \inlinecolorblock{green!55!black}{green!15} ``frozen Kimodo-Root and Kimodo-Body denoisers,'' \inlinecolorblock{blue!55!black}{blue!10} ``trainable ControlNet-Root and ControlNet-Body branches injecting DINOv3~\citep{dinov3} features,'' and \inlinecolorblock{gray!65}{gray!15} ``trainable HandNet denoiser for bilateral hand control.'' Dual-Path Fusion combines motion-query cross-attention with an MLP-based mapping from spatial visual tokens to the future horizon. The predicted root conditions body generation, while future body features condition HandNet; the Motion Projector converts root--body predictions into tracker references. Snowflakes and flames mark frozen and trainable modules, respectively, and dashed boxes indicate Dual-Path Fusion.}
  \label{fig:method_architecture}
\end{figure}

\textbf{Three-Stage Architecture.}
As illustrated in Figure~\ref{fig:method_architecture}, Kimodo-Policy extends Kimodo's~\citep{kimodo} pretrained motion generator into a three-stage root--body--hand architecture.
LLM2Vec~\citep{behnamghader2024llm2veclargelanguagemodels} encodes the instruction, while frozen DINOv3~\citep{dinov3} extracts egocentric visual features.
Root and body ControlNet branches incorporate these visual features through Dual-Path Fusion and inject the resulting residuals into the corresponding frozen denoisers.
The inter-stage dependencies are summarized as
\begin{equation}
\mathbf{z}^{r}_k = \mathcal{G}_{r}(\mathbf{c}_k),
\quad
\mathbf{z}^{b}_k = \mathcal{G}_{b}\!\left(\mathbf{c}_k,\operatorname{sg}(\mathbf{z}^{r}_k)\right),
\quad
\widehat{\mathbf{U}}_k = \mathcal{G}_{h}\!\left(\mathbf{s}^{h}_k,\operatorname{sg}(\mathbf{V}_k),\operatorname{sg}(\mathbf{z}^{b}_k)\right),
\label{eq:three_stage_architecture}
\end{equation}
where $\mathbf{c}_k$ collects the instruction embedding, motion history, and visual features $\mathbf{V}_k$;
$\mathbf{z}^{r}_k$ and $\mathbf{z}^{b}_k$ collectively denote each stage's motion predictions and conditioning features;
and $\operatorname{sg}(\cdot)$ denotes stop-gradient.
Sampling noise is omitted for brevity.
The root stage predicts global translation and heading, whose local representation conditions the body stage to generate articulated motion.
The trainable HandNet then combines future body features, visual features, and the current hand state to predict bilateral hand controls $\widehat{\mathbf{U}}_k$.
All root-to-body conditioning is detached, preventing body-stage gradients from reaching either Kimodo-Root or ControlNet-Root.
Likewise, body and visual features supplied to HandNet are detached, preventing hand supervision from updating Kimodo-Body or the ControlNet branches.
These gradient boundaries separate optimization across the three stages while preserving their forward dependencies for coordinated generation.
Finally, root and body predictions are converted into whole-body tracking references and executed alongside the hand commands.

\textbf{Dual-Path Fusion.} To ground the frozen motion prior in the current scene, we encode the egocentric image into spatial tokens with DINOv3~\citep{dinov3} and process them through trainable root and body ControlNet branches. At selected layers of each Kimodo denoiser, we fuse visual context through two complementary paths. In the cross-attention path, future-motion hidden states serve as queries over the image tokens, allowing each predicted time step to retrieve task-relevant visual evidence. In parallel, a token-axis MLP maps the spatial image-token sequence to the future temporal horizon, providing a direct image-to-time correspondence. We sum the two outputs and pass them through a layer-specific zero-initialized projection to form a residual hint for the corresponding Kimodo layer. These hints are added only to future-motion tokens, leaving the constrained history prefix unchanged. The two paths jointly support content-selective visual retrieval and temporally aligned conditioning.

\textbf{HandNet.} Hand control has different output semantics from root and body motion: it specifies when each hand should open or close, rather than a kinematic pose trajectory. We therefore model bilateral hand control with a dedicated diffusion head. At each denoising step, HandNet jointly processes the current hand state, noisy future hand controls, future body hidden states, and visual tokens, and predicts two hand-control values for every future frame. Conditioning on the predicted body and current scene aligns interaction commands with the evolving whole-body action, while a separate head avoids placing hand labels in the root--body motion space. The head supports binary open/close states and continuous closure changes. During training, its body and visual conditioning features are detached, so hand supervision updates HandNet without sending gradients into the motion denoiser or ControlNet.

\subsection{Adapting Motion Priors for Closed-Loop Control}
\label{sec:t2m_transfer}

\begin{figure}[H]
  \centering
  \includegraphics[width=\textwidth]{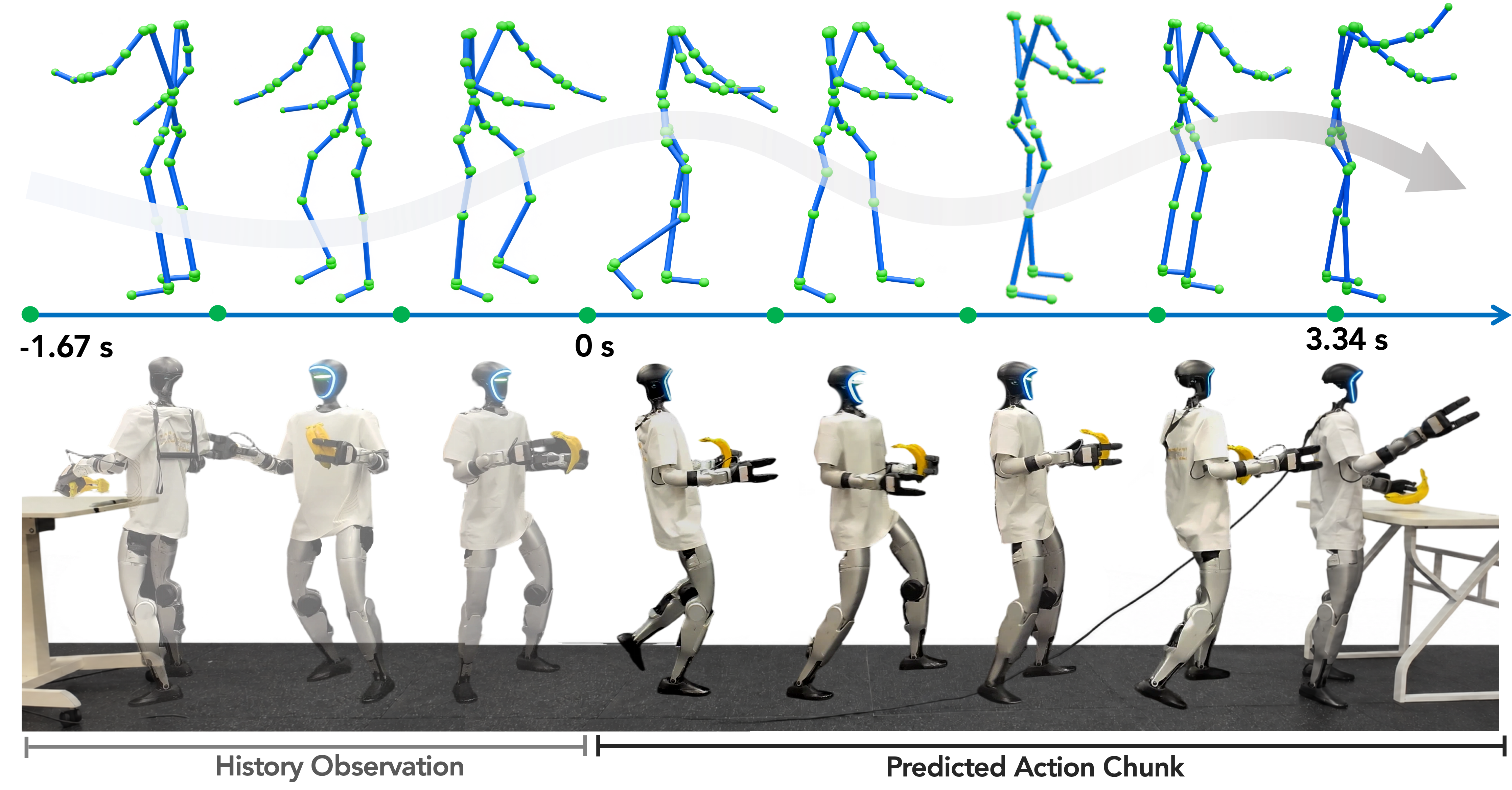}
  \caption{\textbf{From Language-Aligned Motion Priors to Closed-Loop Interaction.} Top: motion generated by the original Kimodo illustrates language-aligned whole-body priors. Bottom: real-world object transfer with Kimodo-Policy across successive closed-loop updates. The time markers and history/prediction brackets are schematic and do not specify single-query durations; actual horizons are detailed in Sec.~\ref{sec:setup_posttraining}. Actual execution repeatedly applies a chunk prefix and replans from updated visual observations and measured motion history.}
  \label{fig:t2m_transfer}
\end{figure}

\textbf{Grounding Language-Aligned Motion Priors.} Kimodo's motion pretraining associates language semantics with coordinated root and body movements~\citep{kimodo}, as illustrated by the sequence generated by the original model in the upper row of Figure~\ref{fig:t2m_transfer}. This language-aligned motion prior provides the generative foundation for Kimodo-Policy. Real-world interaction additionally requires the generated motion to account for the observed scene and the robot's ongoing execution. We learn visual conditioning from robot demonstrations to adapt the inherited motion-generation process to the current scene, while motion-history constraints anchor each prediction to the robot's executed trajectory. The lower row illustrates the resulting closed-loop policy in real-world object transfer. Within this KAM instantiation, policy learning therefore centers on grounding an existing language-aligned motion prior in the environment and execution context, with the pretrained generator retained as the primary action backbone.

\begin{samepage}
\textbf{Motion-History Conditioning.} Applying Kimodo's pretrained sequence generator to closed-loop control requires reconciling coherent motion synthesis with repeated, observation-dependent action selection: the policy must predict a finite action chunk, execute only part of it, and revise subsequent predictions using updated observations. We formulate each policy query as constrained motion completion in Kimodo's kinematic representation. The available measured motion history occupies the prefix of a history--future window, and a feature-wise mask specifies the observed kinematic values. This prefix is restored before each denoising step, while the unconstrained future is generated as the next action chunk. Trainable ControlNet branches inject features from the current egocentric image into the frozen root and body denoisers, grounding predictions in the scene. In the real-world configuration, each query uses up to 100 history frames and predicts 50 future frames at 30~Hz (approximately 1.67~s); a 15-frame prefix (0.5~s) is executed per update (Sec.~\ref{sec:setup_posttraining}). The resulting measured state history and refreshed image then condition the next prediction, allowing subsequent chunks to account for execution deviations and environmental changes. The full interaction in Figure~\ref{fig:t2m_transfer} therefore spans successive prediction--execution updates. This receding-horizon formulation retains the pretrained motion structure while enabling continual visual and state-based correction.\par
\end{samepage}

\subsection{Training and Inference}
\label{sec:training_inference}

\textbf{Training Objective.} We train the ControlNet branches and HandNet with supervised denoising on demonstration windows. For each window, we sample a diffusion timestep and perturb the normalized root--body features and hand targets with Gaussian noise. A feature-wise mask imposes the observed history as a constraint, while supervision is applied to valid future frames. The objective is
\begin{equation}
  \mathcal{L}=\lambda_r\mathcal{L}_{r}+\lambda_b\mathcal{L}_{b}
  +\lambda_h\bigl(\mathcal{L}_{h}+\lambda_{\Delta}\mathcal{L}_{\Delta}\bigr),
  \label{eq:policy_training_objective}
\end{equation}
where $\mathcal{L}_{r}$ and $\mathcal{L}_{b}$ are masked mean squared errors between the predicted clean root and body features, respectively, and their normalized targets, evaluated over valid future frames. For binary control, hand states are encoded in $[-1,1]$; for continuous control, HandNet predicts closure residuals relative to the latest valid observed hand state. The hand reconstruction loss $\mathcal{L}_{h}$ is a masked mean squared error on these targets. The transition loss $\mathcal{L}_{\Delta}$ penalizes the $\ell_1$ error between predicted and demonstrated consecutive hand-control changes, including the transition from the observed state to the first future frame. In continuous mode, predicted residuals are converted back to closure values before computing this loss. Motion losses exclude padded frames, while hand losses additionally mask unavailable labels; transition terms are evaluated only when both adjacent labels are valid. For the main MSE configuration, we use $(\lambda_r,\lambda_b,\lambda_h,\lambda_{\Delta})=(4,1,1,0.2)$. We detach the predicted root before body generation and, in the main configuration, detach root ControlNet features passed to the body branch. HandNet also receives detached body and visual features. These detachments regulate gradient flow between trainable branches, while the Kimodo denoisers and DINOv3~\citep{dinov3} encoder remain frozen.

\textbf{Inference.} At each policy query, we encode the language instruction and current egocentric image, and represent the available proprioceptive history in Kimodo's motion space with frame- and feature-wise observation masks. We initialize the future motion and hand sequences from Gaussian noise and iteratively denoise them with DDIM~\citep{song2022denoisingdiffusionimplicitmodels}. At each step, the observed history is restored, visual features are injected into the frozen root and body denoisers through ControlNet, the predicted root conditions body generation, and HandNet predicts hand controls from the current hand state, visual features, and future body features. After the final denoising step, we invert the motion representation and map the resulting root and joint trajectories, together with bilateral hand commands, to tracker-compatible references. The low-level whole-body tracker executes a prefix of the predicted chunk; the policy then incorporates the updated visual observation and proprioceptive state to replan. This receding-horizon loop combines structured motion generation with continual closed-loop correction.

%% file: content/Experimental_Setup.tex
\section{Experimental Setup}
\label{sec:experimental_setup}

This section describes the data, training protocols, and evaluation environments used to assess Kimodo-Policy. We first detail loco-manipulation policy pretraining on heterogeneous humanoid demonstrations (Sec.~\ref{sec:setup_pretraining}), and then describe post-training and evaluation on the HumanoidArena and SIMPLE simulators as well as on a physical humanoid robot (Sec.~\ref{sec:setup_posttraining}).

\subsection{Pretraining}
\label{sec:setup_pretraining}
\begin{figure}[H]
  \centering
  \includegraphics[width=\textwidth]{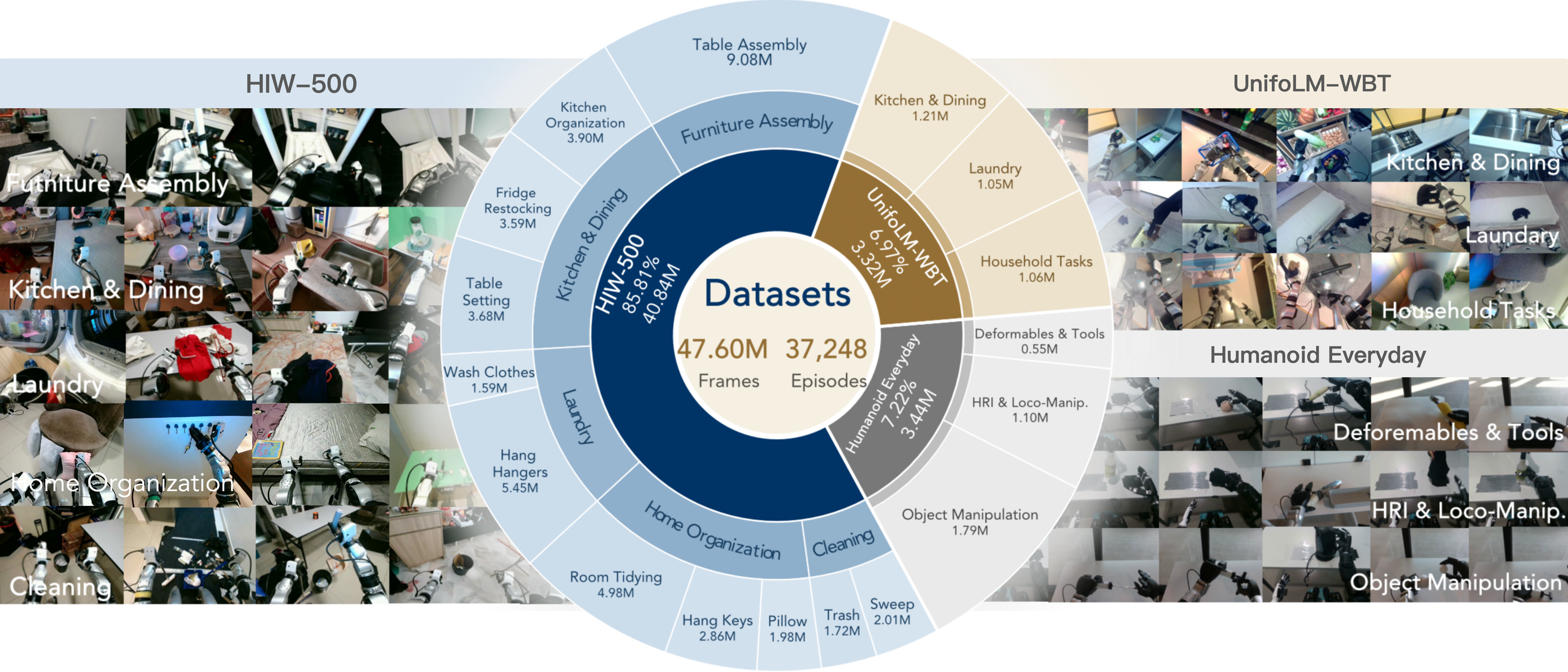}
  \caption{\textbf{Composition of the Policy Pretraining Corpus.} The corpus combines HIW-500, 14 task subsets from UniFoLM-WBT available when the corpus was assembled, and Humanoid Everyday. UniFoLM-WBT entries released after corpus construction are excluded.}
  \label{fig:pretraining_datasets}
\end{figure}

\textbf{Dataset.} Here, policy pretraining refers to training the visual conditioning modules and hand head on robot demonstrations while retaining the frozen, pretrained Kimodo motion backbone. We assemble a 419-hour pretraining corpus from three publicly released humanoid datasets (Fig.~\ref{fig:pretraining_datasets}). HIW-500 provides teleoperated Unitree G1 demonstrations collected in real homes. From UniFoLM-WBT, we use the 14 task-specific G1 subsets available when our corpus was assembled: the BrainCo subsets \emph{Collect Plates into Dishwasher}, \emph{Make the Bed}, \emph{Pick Up Medicine}, and \emph{Pickup Pillow}; the Dex1 subset \emph{Put Clothes into Washing Machine}; and the Inspire subsets \emph{Clean the Living Room}, \emph{Collect Clothes (MainCamOnly)}, \emph{Pick Up Drinks}, \emph{Pickup Pillow (MainCamOnly)}, \emph{Put Clothes into Basket}, \emph{Put Clothes into Washing Machine}, \emph{Put Clothes into Washing Machine (MainCamOnly)}, \emph{Put Drinks into Fridge}, and \emph{Put Vegetables into Basket}. Main-camera-only variants are included when available, while UniFoLM-WBT entries released after this snapshot are excluded. Humanoid Everyday~\citep{humanoid_everyday} contributes 10.3K trajectories, over three million frames, and 260 tasks spanning dexterous manipulation, articulated-object interaction, human--robot interaction, and locomotion-integrated behaviors. We form training examples from an egocentric image, a language instruction, a 100-frame motion history, and a 50-frame future action chunk sampled at 30~Hz, converting trajectories to Kimodo's motion representation.

\textbf{Implementation Details.} We pretrain the policy for 1,000k optimization steps on 16 NVIDIA H200 GPUs, taking approximately 60 hours with bfloat16 mixed precision and a global batch size of 1024. During policy pretraining, hand supervision is represented as a two-dimensional binary vector indicating the open/closed state of the left and right hands. This shared representation accommodates the different hand morphologies, degrees of freedom, and control conventions in the source datasets, preserving grasp-and-release semantics while providing a consistent target for the hand head. We use AdamW with learning rates of $3\times10^{-5}$ for the ControlNet Transformer layers and $1\times10^{-4}$ for the ControlNet conditioning modules and hand head, together with a weight decay of $0.01$. The learning rate follows a cosine schedule with 1\% warm-up and a minimum of 10\% of its initial value.

\subsection{Post-Training}
\label{sec:setup_posttraining}
\textbf{HumanoidArena Benchmark.} HumanoidArena~\citep{humanoidarena,sonic} provides seven leg-critical tasks: three human--object and four human--scene interaction tasks requiring coordinated locomotion, balance, posture adjustment, and manipulation. Its SONIC data contain 100 successful demonstrations per task (700 total) with synchronized egocentric RGB, proprioceptive states, and whole-body targets. We use all seven tasks for single-task fine-tuning and joint multi-task training, the latter being more challenging because one policy must acquire all tasks simultaneously. We evaluate closed-loop performance with SONIC using success rate (SR) and average fall rate (AFR), capturing both task completion and stability during whole-body task execution.

\textbf{SIMPLE Benchmark.} SIMPLE~\citep{simple} integrates MuJoCo contact dynamics with Isaac Sim rendering for scalable evaluation of humanoid loco-manipulation. From the released task suite, we select eight for evaluation: Handover, Mobile P\&P, XMoveBendPick, CloseDoor, OpenOven, OpenFaucet, Push Office Chair, and OpenTrashCan. Together, these tasks cover object handover, mobile pick-and-place, articulated-object interaction, and household-object manipulation, allowing us to assess policy robustness and execution stability in contact-rich whole-body control.

\begin{figure}[H]
  \centering
  \includegraphics[width=\textwidth]{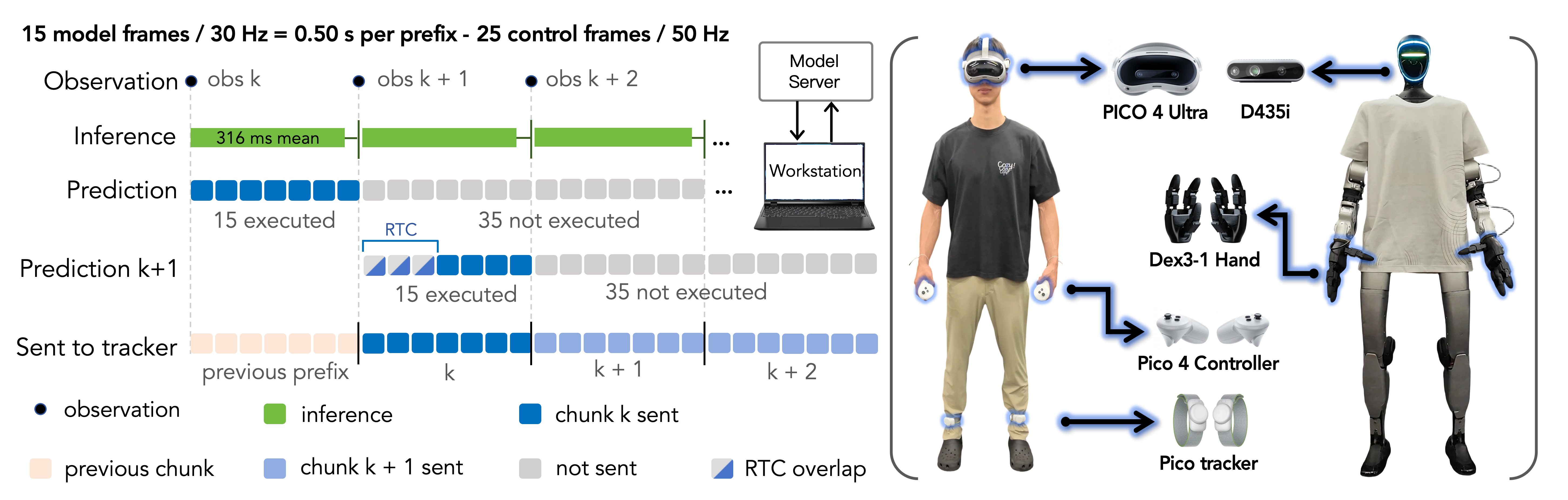}
  \caption{\textbf{Real-World Teleoperation and Deployment Setup.} Left: real-time action-chunk execution, where each policy update supplies a 15-frame prefix at 30\,Hz to the 50\,Hz tracker, with RTC overlap between consecutive chunks. Right: PICO-based whole-body teleoperation and the humanoid platform equipped with a RealSense D435i camera and Dex3-1 hands.}
  \label{fig:real_world_setup}
\end{figure}

\begin{figure}[H]
  \centering
  \includegraphics[width=\textwidth]{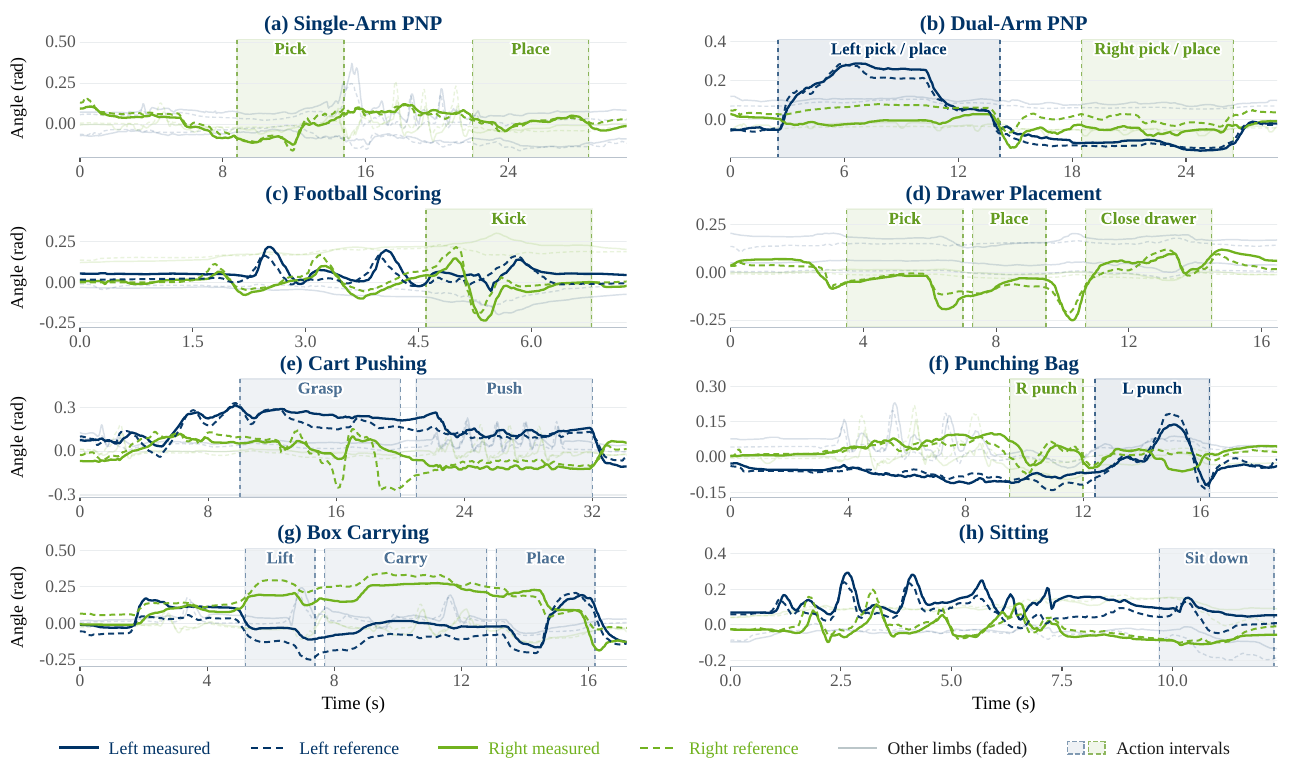}
  \caption{\textbf{Motion Profiles of the Collected Real-World Demonstrations.} One episode per task is selected to illustrate pronounced task-related motion. Curves show mean angles over six joints per leg or seven per arm. Blue and green denote left and right limbs; solid and dashed curves indicate measured states and recorded motion references, respectively. Task-relevant limbs are emphasized, with other limbs faded. Shaded regions mark the labeled action intervals, and vertical dashed lines delimit their approximate boundaries.}
  \label{fig:real_world_data_profiles}
\end{figure}

\textbf{Real-World Data Collection and Setup.} We collect real-world demonstrations with a teleoperation and data-recording system extended from NVIDIA GR00T-WholeBodyControl~\citep{sonic}. As shown on the right of Fig.~\ref{fig:real_world_setup}, the operator uses a PICO 4 Ultra headset and hand controllers together with PICO body trackers for whole-body and lower-limb tracking. The humanoid is equipped with an Intel RealSense D435i camera and Dex3-1 dexterous hands. We define eight real-world tasks spanning locomotion, balance, and single- and dual-arm manipulation, and record 25--30 demonstrations per task. Figure~\ref{fig:real_world_data_profiles} visualizes the collected demonstrations across all eight tasks by comparing measured limb-joint trajectories with their recorded motion references. The shaded intervals mark key action phases, such as kicking, object placement, and sitting, making the task-dependent temporal structure of the demonstrations explicit. We store the demonstrations in a LeRobot v2.1~\citep{cadene2026lerobotopensourcelibraryendtoend}-compatible format, extended with Kimodo-specific motion and control fields for policy training with a motion prior.

\textbf{Implementation Details.} We fine-tune the Kimodo-Policy variants with bfloat16 precision on four GPUs using 30\,Hz motion sequences, a 100-frame history, and a 50-frame prediction horizon. HumanoidArena single-task policies use a global batch size of 64 for 200k steps, while the joint seven-task policy uses a batch size of 128 for 500k steps; SIMPLE uses a batch size of 64 for 200k steps, and real-world policies use a batch size of 32 for 50k steps. Variants without and with policy pretraining share the same fine-tuning setup and retain the same pretrained Kimodo motion backbone; they differ in the initialization of the trainable conditioning modules and hand head. The four-layer ControlNet detaches the root-control input to body prediction. We use the objective in Eq.~\ref{eq:policy_training_objective}, with weights 4, 1, 1, and 0.2 for root motion, body motion, hand control, and hand transitions, respectively; reconstruction terms use MSE and the hand-transition term uses an $\ell_1$ loss. Optimization uses AdamW (weight decay 0.01) and a cosine schedule with 1\% warm-up and a 10\% minimum learning-rate ratio. The ControlNet Transformer layers, adapters, and hand head use learning rates of $3\times10^{-5}$, $1\times10^{-4}$, and $1\times10^{-4}$ for HumanoidArena fine-tuning, and $1\times10^{-5}$, $3\times10^{-5}$, and $3\times10^{-5}$ for SIMPLE and real-world fine-tuning. SIMPLE training uses color jitter and continuous hand-control targets; real-world training uses color jitter and random crops, with binary hand commands conditioned on measured hand closure. At inference, Kimodo-Policy runs on one NVIDIA RTX 4090 with 10 diffusion denoising steps and predicts a 50-frame action horizon at 30\,Hz. We execute 15 predicted frames per policy update on HumanoidArena and the real robot, and either 15 or 50 frames on SIMPLE depending on the evaluation setting. The SONIC tracker runs at 50\,Hz; real-time chunking (RTC)~\citep{rtc} is enabled only for real-world deployment and disabled for HumanoidArena and SIMPLE.

%% file: content/Experiments_Results.tex
\section{Experimental Results}
\label{sec:experiments}

We evaluate motion generation as a foundation for humanoid control through benchmark and real-world experiments (Secs.~\ref{sec:exp_simulation}--\ref{sec:exp_real}), followed by analyses of motion priors, scaling, and adaptation design (Secs.~\ref{sec:exp_t2m}--\ref{sec:exp_ablation}). Unless explicitly ablated, all Kimodo-Policy variants retain the pretrained motion backbone. Kimodo-Policy denotes direct downstream fine-tuning, while Kimodo-Policy$^{*}$ additionally initializes the conditioning modules and hand head through policy pretraining on 419 hours of robot-interaction data.

\subsection{Evaluation on the HumanoidArena Benchmark}
\label{sec:exp_simulation}

\begin{figure}[H]
  \centering
  \includegraphics[width=\textwidth]{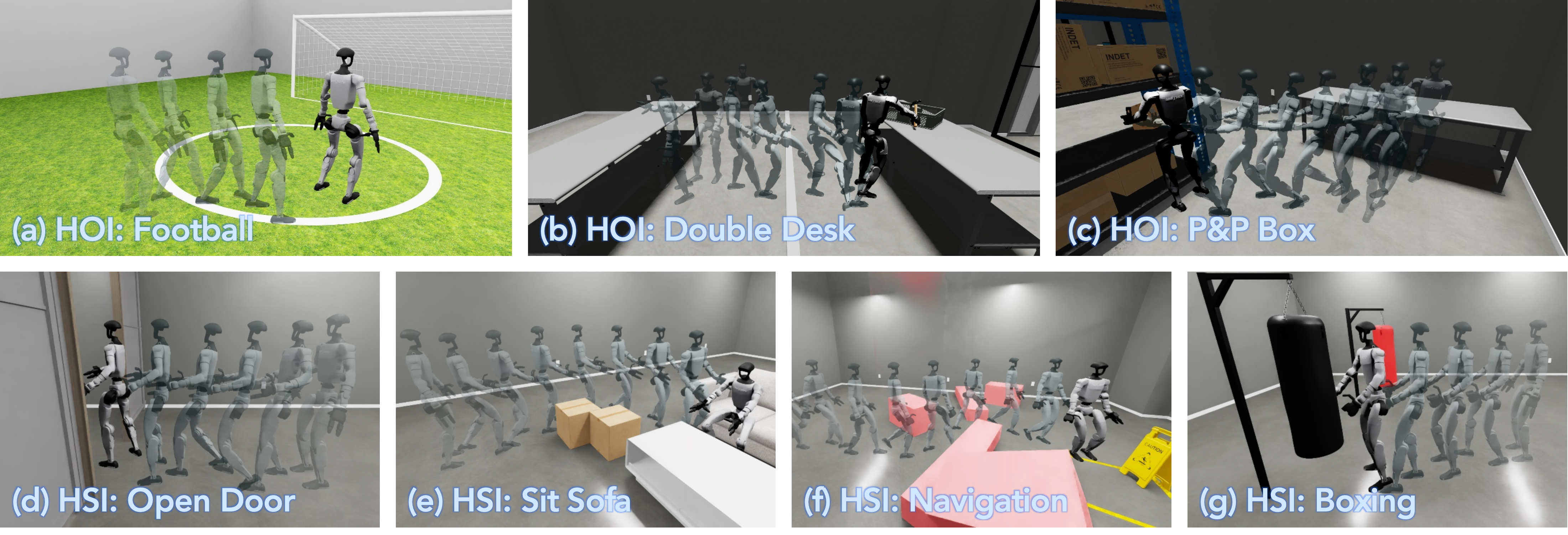}
  \caption{\textbf{Visualization of the Seven HumanoidArena Tasks.} The benchmark includes Football, DoubleDesk, P\&Pbox, OpenDoor, SitSofa, Boxing, and VisNavi, covering language-conditioned humanoid behavior across human--object and human--scene interaction with coordinated locomotion, balance, and manipulation.}
  \label{fig:humanoidarena}
\end{figure}

HumanoidArena~\citep{humanoidarena} comprises seven language-conditioned tasks spanning human--object interaction (HOI) and human--scene interaction (HSI), with substantial requirements for locomotion, posture adjustment, and manipulation (Fig.~\ref{fig:humanoidarena}). We evaluate both independent task-specific policies and a single policy jointly trained on all seven tasks. Success rate (SR) measures task completion, while average fall rate (AFR) characterizes execution stability within the SONIC-based control stack~\citep{sonic}. Real-time chunking (RTC) is disabled in our simulation evaluations.

\textbf{Single-task results.} Table~\ref{tab:arena_single_task} shows that Kimodo-Policy achieves the highest overall SR of 69.76\%, exceeding Diffusion Policy (DP)~\citep{chi2023diffusionpolicy}, the strongest baseline in this setting, by 9.76 percentage points. Both Kimodo-Policy variants obtain an AFR of 4.29\%, the lowest reported in the table. The aggregate advantage is concentrated in HSI: direct fine-tuning achieves an HSI average of 84.58\%, compared with 65.83\% for DP, including 100.0\% on SitSofa and 75.0\% on VisNavi. In contrast, its HOI average of 50.00\% remains below DP's 52.22\%. Thus, the motion-based policy is particularly effective on the evaluated scene-interaction tasks, while its advantages do not extend uniformly to every interaction category. Additional policy pretraining yields 67.14\% overall SR, below the 69.76\% obtained by direct fine-tuning. It improves OpenDoor from 90.0\% to 93.3\% and VisNavi from 75.0\% to 76.7\%, but reduces performance on the remaining tasks. These results show that the pretrained motion backbone can support effective task-specific adaptation without the additional robot-interaction pretraining stage; they also indicate that the benefit of that stage depends on the downstream training setting. These scene-interaction tasks require coordinated changes in posture and global position, making the strong HSI performance and low AFR consistent with transferring useful whole-body movement structure.

\input{table/arena_single_task}

\textbf{Multi-task results.} Table~\ref{tab:arena_multi_task} extends the comparison to a shared seven-task policy and includes additional VLA and WAM baselines. Kimodo-Policy$^{*}$ achieves the highest overall SR among the compared methods at 72.14\%, exceeding GR00T N1.7 (69.70\%) and GigaWorld-Policy (68.39\%). The gain over GR00T N1.7, the strongest baseline in the table, is 2.44 percentage points. Its AFR is approximately 4.3\%, the lowest among methods for which this metric is reported; AFR is unavailable for several of the added baselines. The task breakdown clarifies the source of this aggregate advantage. Kimodo-Policy$^{*}$ achieves the highest HSI average, approximately 84.2\%, and leads on P\&Pbox and VisNavi while matching HuGe-GR00T on OpenDoor and the directly fine-tuned Kimodo-Policy on SitSofa. Its HOI average, approximately 56.1\%, is below GR00T N1.7's 62.22\% and GigaWorld-Policy's 60.14\%. In particular, Football and DoubleDesk remain challenging relative to the strongest task-specific competitors. The results therefore support the effectiveness of this KAM instantiation across a diverse task set, with its strongest aggregate advantage in scene interaction. Within our framework, policy pretraining increases overall multi-task SR from 67.63\% to 72.14\%, a gain of 4.51 percentage points, with improvements on five tasks, unchanged performance on SitSofa, and a decrease on DoubleDesk. Its benefit in joint training contrasts with the single-task results, suggesting that interaction pretraining is more useful when one set of conditioning modules must support multiple behaviors.

\input{table/arena_multi_task}

\begin{figure}[H]
  \centering
  \includegraphics[width=\textwidth]{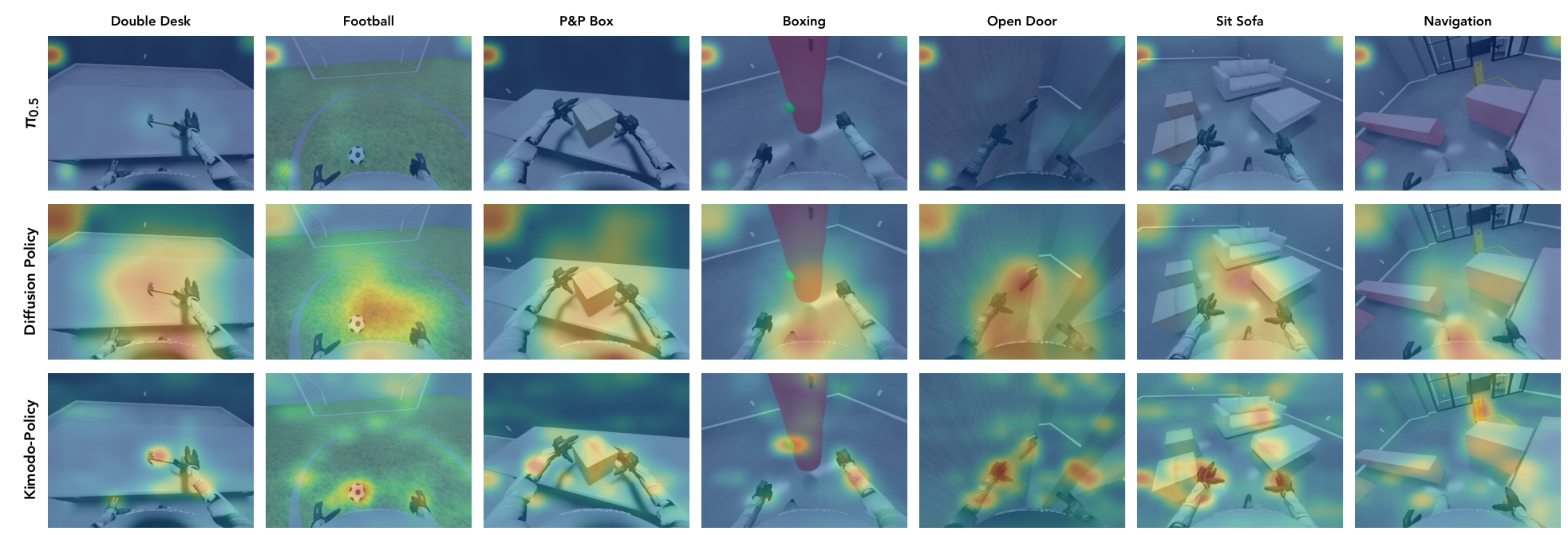}
  \caption{\textbf{Qualitative Visual Responses Across HumanoidArena Tasks.} Columns show DoubleDesk, Football, P\&Pbox, Boxing, OpenDoor, SitSofa, and VisNavi; rows show $\pi_{0.5}$, Diffusion Policy, and Kimodo-Policy. Kimodo-Policy exhibits localized responses near interaction-relevant regions in the displayed examples.}
  \label{fig:attention}
\end{figure}

\textbf{Qualitative visual grounding.} Figure~\ref{fig:attention} visualizes responses around task-relevant objects and interaction regions, including the ball, box, and punching bag. In the displayed examples, Kimodo-Policy exhibits localized responses near these regions, whereas DP often produces more spatially diffuse activation. This pattern is consistent with the intended role of perception in a KAM: visual observations condition where the inherited movement patterns should be expressed. These maps provide qualitative context for the control results, but do not establish causal feature importance or instruction sensitivity, since the examples also differ in scene content.

\subsection{Evaluation on the SIMPLE Benchmark}
\label{sec:exp_simple}

\begin{figure}[H]
  \centering
  \includegraphics[width=\textwidth]{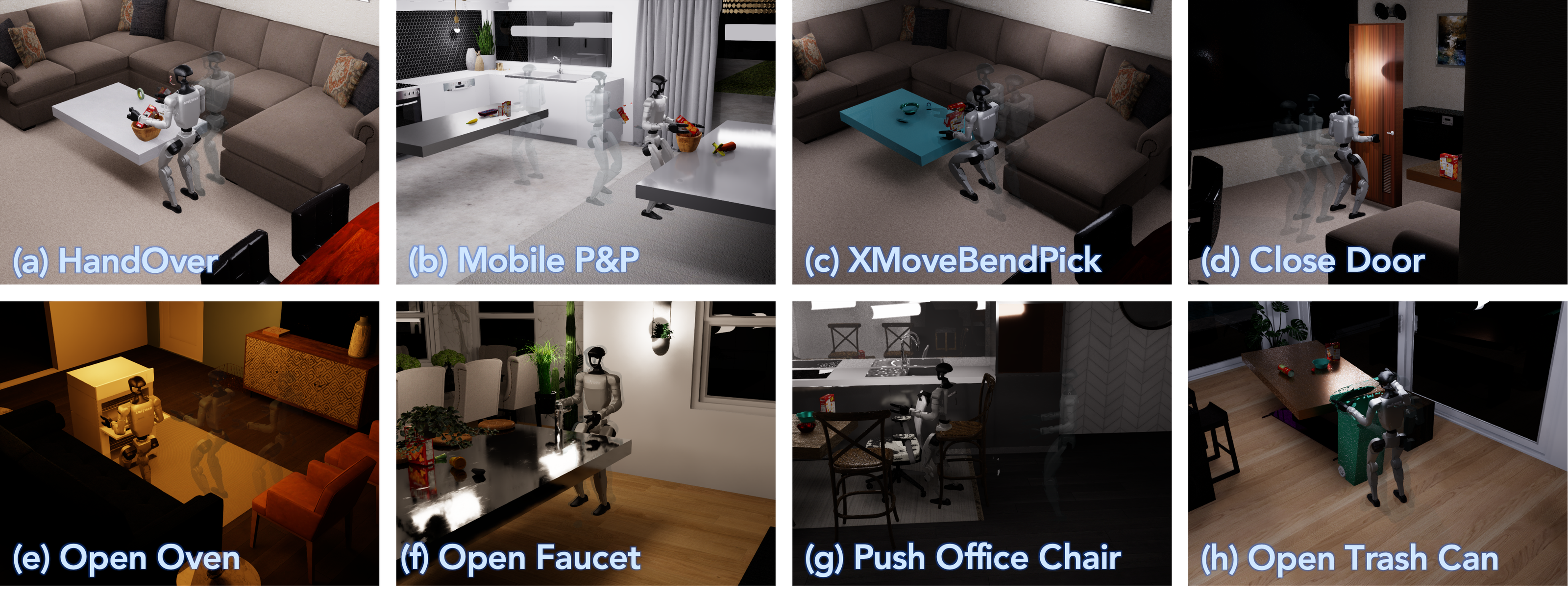}
  \caption{\textbf{Evaluated SIMPLE Task Suite.} The selected tasks cover object handover and mobile pick-and-place, locomotion-integrated object pickup, articulated-object interaction, and household-object manipulation. Panels show (a) Handover, (b) Mobile P\&P, (c) XMoveBendPick, (d) CloseDoor, (e) OpenOven, (f) OpenFaucet, (g) Push Office Chair, and (h) OpenTrashCan.}
  \label{fig:simple_tasks}
\end{figure}

We next evaluate eight SIMPLE~\citep{simple} tasks covering handover, mobile manipulation, articulated-object interaction, and object pushing (Fig.~\ref{fig:simple_tasks}). An independent Kimodo-Policy$^{*}$ is fine-tuned for each task and evaluated at the L0--L2 randomization levels, which introduce progressively stronger changes in appearance, lighting, and target pose. Table~\ref{table:simple_main_results} reports ten trials per level and task, with RTC disabled for our policies. This evaluation tests whether the motion-based action generator can be adapted to different interaction requirements and remain effective under the prescribed perturbations.

\input{table/simple_single_task}

\textbf{Single-task results.} Kimodo-Policy$^{*}$ achieves 75.83\% mean SR across the eight tasks, compared with 75.42\% for $\Psi_0$~\citep{psi0}, the strongest baseline with complete task coverage. These values correspond to 182 and 181 successful trials out of 240, respectively. Baseline results are compiled from the official SIMPLE release and a subsequent evaluation~\citep{simple,chen2026memorybeforegoal}, as specified in the table caption; they are not a unified reevaluation under our training and execution pipeline. The aggregate performance is therefore comparable, with a small numerical advantage for Kimodo-Policy$^{*}$. Kimodo-Policy$^{*}$ attains the best or joint-best SR on five tasks. It leads on Handover (25/30 successes), Mobile P\&P (21/30), and OpenFaucet (16/30), compared with the strongest baseline counts of 24/30, 18/30, and 11/30, respectively. It also achieves 30/30 on both CloseDoor and Push Office Chair, tying the best reported results. It ranks second on the remaining three tasks: XMoveBendPick (24/30 versus 28/30 for $\Psi_0$), OpenOven (17/30 versus 22/30 for InternVLA-M1~\citep{internvla_m1}), and OpenTrashCan (19/30 versus 25/30 for $\Psi_0$). This distribution demonstrates broad applicability of the motion-based policy while identifying interaction types for which other approaches remain stronger. Averaged across the eight tasks, Kimodo-Policy$^{*}$ achieves 87.50\%, 71.25\%, and 68.75\% SR at L0, L1, and L2, respectively. Performance remains perfect on CloseDoor and Push Office Chair, whereas Mobile P\&P declines from 9/10 at L0 to 4/10 at L2. The adapted policy therefore remains sensitive to perceptual and geometric variation despite its inherited motion prior. Together, these results support competitive task performance while identifying robustness as an important remaining challenge.

\subsection{Evaluation on the Real-World Humanoid Robot}
\label{sec:exp_real}

\begin{figure}[H]
  \centering
  \includegraphics[width=\textwidth]{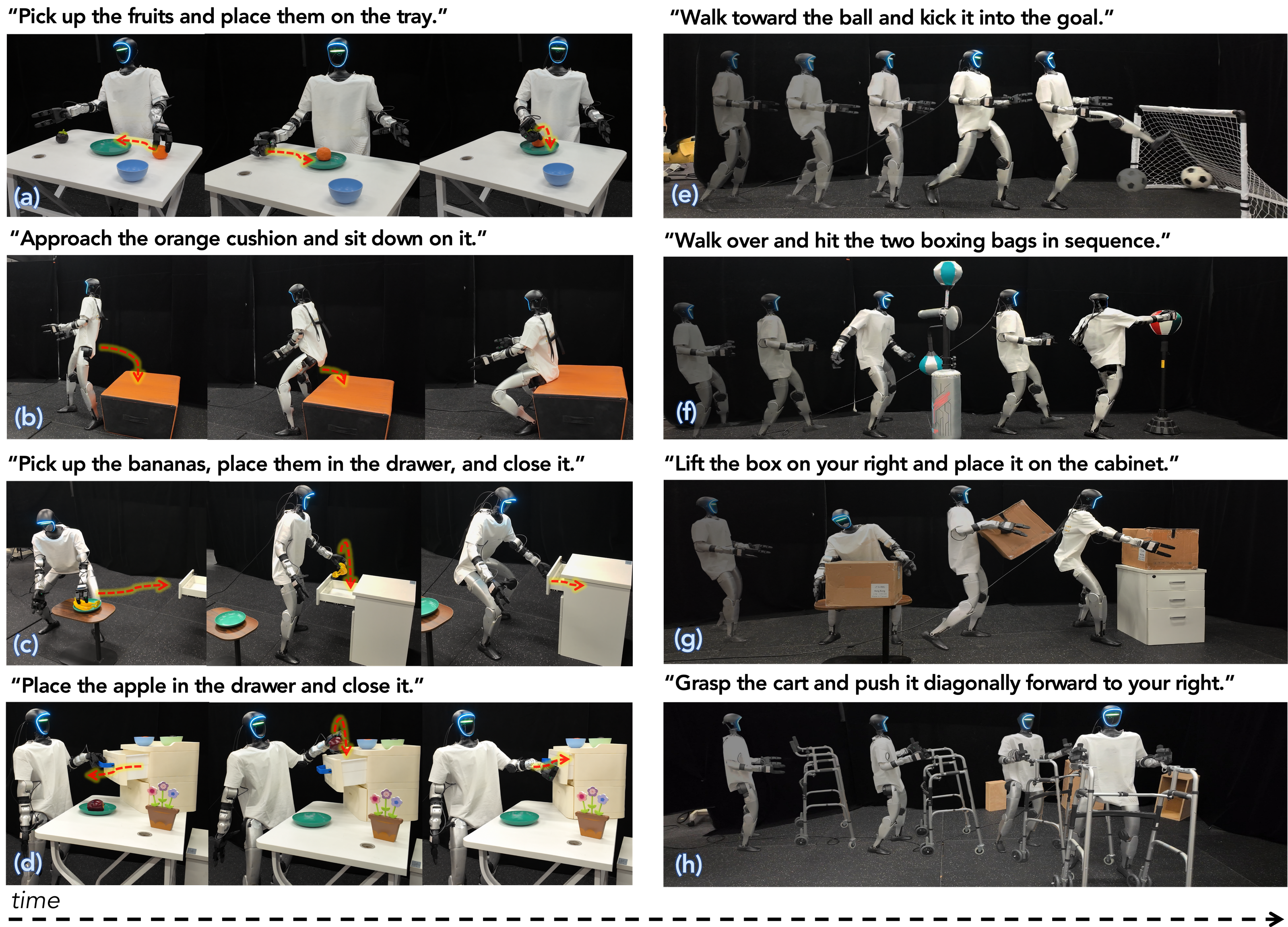}
  \caption{\textbf{Examples of Real-World Humanoid Manipulation.} The image shows representative interaction scenarios involving articulated objects, object transport, and coordinated whole-body motion.}
  \label{fig:real_world}
\end{figure}

We deploy Kimodo-Policy on a physical humanoid to evaluate the complete perception--generation--tracking pipeline beyond simulation. The eight tasks require different combinations of locomotion, posture adjustment, single- and dual-arm coordination, and object interaction (Fig.~\ref{fig:real_world}). Each policy is evaluated for ten trials per task. Deployment follows Sec.~\ref{sec:setup_posttraining}, including RTC for Kimodo-Policy. Both Kimodo-Policy variants retain the pretrained motion backbone and differ in whether the trainable modules undergo policy pretraining before task-specific fine-tuning.

\begin{figure}[H]
  \centering
  \includegraphics[width=\textwidth]{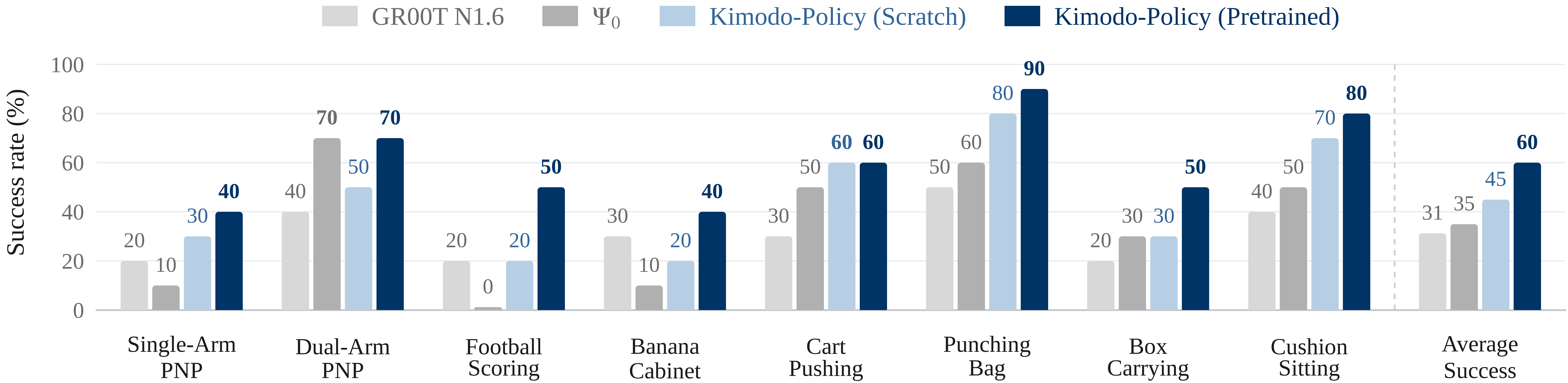}
  \caption{\textbf{Real-World Task Success.} Success rates for GR00T N1.6, $\Psi_0$, and Kimodo-Policy with and without additional robot-interaction policy pretraining. Both Kimodo-Policy variants retain the pretrained motion backbone. Each task is evaluated for ten trials; the rightmost group reports the mean across eight tasks. The plot label \emph{Scratch} denotes the variant without additional policy pretraining, while \emph{Pretrained} denotes the variant with it.}
  \label{fig:real_world_success_rates}
\end{figure}

\textbf{Real-world results.} As shown in Fig.~\ref{fig:real_world_success_rates}, Kimodo-Policy without additional policy pretraining achieves 45\% mean SR, compared with approximately 31\% for GR00T N1.6~\citep{grootn1} and 35\% for $\Psi_0$. Policy pretraining increases the mean to 60\%, corresponding to 48 successful trials out of 80 and a 15-percentage-point gain over direct fine-tuning. The pretrained variant improves on seven tasks and matches the directly fine-tuned variant on Cart Pushing (60\%). Its highest SRs are 90\% on Punching Bag, 80\% on Cushion Sitting, and 70\% on Dual-Arm PNP. Improvements are largest on Football Scoring (20\% to 50\%), followed by Dual-Arm PNP, Banana Cabinet, and Box Carrying (20 percentage points each). These results provide initial evidence for KAMs on a physical robot. A motion generator conditioned on current observations and coupled to a whole-body tracker supports the evaluated locomotion and manipulation tasks. Because the motion backbone is shared by both variants, their performance difference supports the complementary value of robot-interaction pretraining for adapting the conditioning modules and hand head. The baseline comparison evaluates the deployed systems as a whole; the effect of RTC is not separately ablated. With ten trials per task, the reported rates characterize performance on this evaluation set rather than establish deployment reliability across unrestricted environments.

\subsection{How Does the Choice of Motion Prior Affect Downstream Control?}
\label{sec:exp_t2m}

A central implication of KAMs is that the action-generating backbone should matter beyond its parameter count. We examine this by substituting two pretrained Kimodo-G1 variants~\citep{kimodo} while retaining the policy architecture, adaptation procedure, and HumanoidArena evaluation protocol. Both backbones contain 282M parameters. The released SEED-v1 and RP-v1 checkpoints specify motion corpora of 288 h and approximately 560 h, respectively, with 90\%/10\% train/test splits. This comparison evaluates the downstream consequences of selecting different motion priors at matched backbone capacity.

\input{table/proof_t2m}

\textbf{Motion-prior transfer results.} Table~\ref{tab:t2m_dependency} reports an overall SR of 64.29\% for SEED-v1 and 67.63\% for RP-v1, a difference of 3.34 percentage points. The gains are task dependent: RP-v1 improves DoubleDesk, P\&Pbox, OpenDoor, and SitSofa, with the largest increase on OpenDoor (60.0\% to 93.3\%). Conversely, Football declines from 50.0\% to 35.0\%, Boxing from 71.67\% to 66.7\%, and VisNavi from 70.0\% to 61.7\%. The overall improvement is thus driven primarily by the net gain on scene-interaction tasks, rather than a uniform improvement across behaviors. The fixed parameter budget makes this result relevant to the KAM premise: the choice of pretrained motion generator affects downstream control even when policy capacity is unchanged. However, the two corpora differ in both size and content, and Table~\ref{tab:t2m_dependency} does not report an independent motion-generation quality metric. The comparison therefore supports the importance of the inherited motion prior, but does not establish a monotonic relationship between T2M quality and control performance or isolate motion-data scale as the cause of the gain.

\subsection{Scaling Robot-Interaction Data and Trainable Policy Capacity}
\label{sec:exp_scaling}

We examine two ways of strengthening the adaptation of a fixed motion backbone: increasing robot-interaction pretraining data and increasing trainable policy capacity. The data sweep holds the policy architecture and downstream seven-task adaptation protocol fixed. The capacity sweep retains the same pretrained motion backbone, downstream data, and optimization protocol while varying the trainable parameter budget. RTC is disabled throughout. These experiments concern policy adaptation around the motion prior, rather than scaling the motion generator itself.

\begin{figure}[H]
  \centering
  \includegraphics[width=\textwidth]{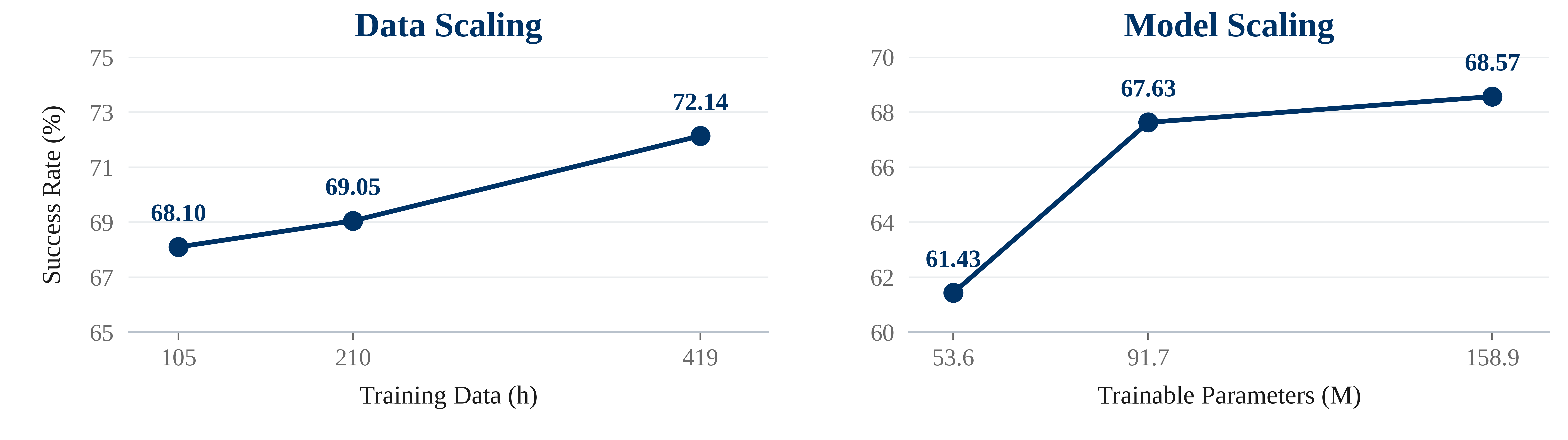}
  \caption{\textbf{Scaling Policy Adaptation Around a Fixed Motion Backbone.} Overall success on the seven HumanoidArena tasks as a function of robot-interaction policy-pretraining data (left) and trainable policy parameters (right). The motion backbone is held fixed in both sweeps. Each panel follows its respective controlled adaptation setting and should not be interpreted as a joint data--capacity sweep.}
  \label{fig:scaling_analysis}
\end{figure}

\textbf{Scaling results.} Figure~\ref{fig:scaling_analysis} shows that overall SR increases from 68.10\% with 105 h of policy-pretraining data to 69.05\% with 210 h and 72.14\% with 419 h, a total gain of 4.04 percentage points. With the motion backbone fixed, these gains show that additional interaction demonstrations can improve the learned conditioning and hand-control components. This complements the role of motion pretraining in the KAM formulation: an inherited movement repertoire provides a foundation, while robot-interaction data adapt its use to observations and task objectives. The observed increase is consistent across the three evaluated data budgets, although it does not by itself establish a scaling law or predict gains beyond this range. For trainable policy capacity, increasing the trainable parameter count from 53.6M to 91.7M raises overall SR from 61.43\% to 67.63\%; a further increase to 158.9M yields 68.57\%. The total improvement is 7.14 percentage points, with a larger gain in the first increment (6.20 points) than in the second (0.94 points). Thus, downstream performance benefits from additional adaptation capacity within the tested range, with diminishing observed returns. The 91.7M configuration used in the main policy remains within 0.94 points of the largest evaluated configuration while using approximately 42\% fewer trainable parameters. This comparison supports the selected trainable parameter budget; inference efficiency is not measured in this sweep.

\begingroup
\titleformat{\subsection}
  {\normalfont\fontsize{14pt}{17pt}\selectfont\bfseries\color{black}}
  {\thesubsection}{0.55em}{}
\subsection{Ablation Study}
\label{sec:exp_ablation}
\color{black}

We distinguish motion-backbone pretraining from robot-interaction policy pretraining. All variants are trained jointly on the seven HumanoidArena tasks; Fig.~\ref{fig:training_loss_comparison} reports mean SR over the three HOI tasks, DoubleDesk, Football, and P\&Pbox. The reference achieves 51.67\% with a pretrained, frozen Kimodo backbone and downstream-trained conditioning modules and HandNet, without additional policy pretraining. \emph{Random initialization} (w/o T2M) retains the same Kimodo root--body architecture but randomly initializes and trains its weights. \emph{Unfrozen backbone} (w/o Frozen) instead initializes that network from the pretrained checkpoint and fine-tunes it. \emph{Original Kimodo objective} (w/o MSE) replaces the root--body MSE objective with component-weighted Smooth L1 reconstruction and a forward-kinematics consistency term. \emph{Without detachment} (w/o Detach) allows body-stage gradients to update the root ControlNet. Each component variant is compared with the reference, while Kimodo-Policy$^{*}$ separately evaluates additional policy pretraining. These results characterize HOI performance and are distinct from the seven-task overall scores in Table~\ref{tab:arena_multi_task}.

\begin{figure}[!t]
  \centering
  \includegraphics[width=\textwidth]{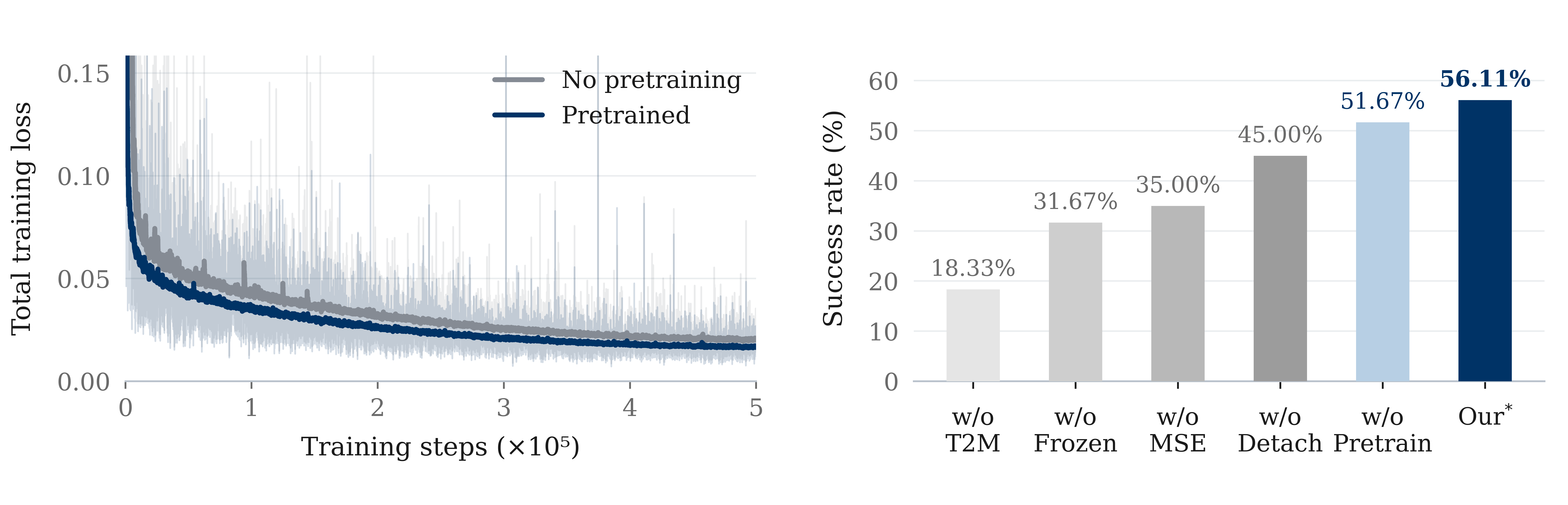}
  \caption{\textbf{Policy Adaptation Dynamics and Ablations.} Left: downstream training-loss trajectories with and without robot-interaction policy pretraining; both retain the pretrained motion backbone. Right: mean SR on the three HOI tasks after joint seven-task training. Relative to the reference without policy pretraining (light blue), gray bars show random initialization and training of the same Kimodo network (w/o T2M), fine-tuning the pretrained backbone (w/o Frozen), the original Kimodo motion objective (w/o MSE), and removal of root--body gradient detachment (w/o Detach). Dark blue adds robot-interaction policy pretraining. The ablation scores are HOI averages, not seven-task overall SR.}
  \label{fig:training_loss_comparison}
\end{figure}

\textbf{Ablation results.} Random initialization yields 18.33\% SR, compared with 31.67\% when the pretrained backbone is fine-tuned and 51.67\% when it remains frozen. Among the two variants with a trainable motion backbone, pretrained initialization improves SR by 13.34 percentage points, supporting the value of inherited motion priors in the evaluated setting. Keeping the pretrained backbone frozen provides a further 20.00-point advantage over fine-tuning it, supporting adaptation through separate conditioning modules. Replacing the root--body MSE objective with the original Kimodo objective reduces SR to 35.00\%, a 16.67-point drop; this comparison favors the selected reconstruction objective and weighting for downstream policy adaptation. \mbox{Removing} root--body gradient detachment yields 45.00\%, 6.67 points below the reference. Thus, retaining the forward dependence of body generation on root predictions while separating their gradient updates improves HOI performance in this setting. Finally, robot-interaction policy pretraining raises SR from 51.67\% to 56.11\%, a gain of 4.44 points, and generally lowers downstream training loss. Since both variants retain the same pretrained, frozen motion backbone, this gain supports the complementary role of initializing the conditioning modules and hand head from interaction demonstrations. Together, these HOI results support pretrained motion initialization, frozen-backbone adaptation, and additional interaction pretraining as distinct contributors to the evaluated Kimodo-Policy configuration.

\endgroup

%% file: table/arena_single_task.tex
\begin{table}[H]
  \caption{
    \textbf{\textit{Single-task training results on HumanoidArena.}} Each policy is fine-tuned independently on one task. ACT, DP, FM, and $\pi_{0.5}$ results are taken from the official HumanoidArena report~\citep{humanoidarena}. Values are mean $\pm$ standard deviation; $^{*}$ denotes additional robot-interaction policy pretraining; both Kimodo-Policy variants retain the pretrained motion backbone. Bold and underlined entries indicate the best and second-best results, respectively.
   Baselines include ACT~\citep{ACT}, DP~\citep{chi2023diffusionpolicy}, FM~\citep{FM_objective}, $\pi_{0.5}$~\citep{pi05}.
  }
  \label{tab:arena_single_task}
  \centering
  \arrayrulecolor{hclblue}
  \begin{adjustbox}{max width=\textwidth}
  \setlength{\tabcolsep}{4.2pt}
  \renewcommand{\arraystretch}{1.08}

  \begin{tabular}{cccccc|cccccc}
  \toprule
  & &
  \multicolumn{4}{c|}{\textbf{Human--Object Interaction}} &
  \multicolumn{5}{c}{\textbf{Human--Scene Interaction}} & \\
  \cmidrule(lr){3-6}
  \cmidrule(lr){7-11}

  \multirow{-2}{*}{\textbf{Method}} &
  \multirow{-2}{*}{\textbf{AFR~($\downarrow$)}} &
  \textbf{\textsc{Football}} &
  \textbf{\textsc{DoubleDesk}} &
  \textbf{\textsc{P\&Pbox}} &
  \textbf{\textsc{Avg~($\uparrow$)}} &
  \textbf{\textsc{OpenDoor}} &
  \textbf{\textsc{SitSofa}} &
  \textbf{\textsc{Boxing}} &
  \textbf{\textsc{VisNavi}} &
  \textbf{\textsc{Avg~($\uparrow$)}} &
  \multirow{-2}{*}{\textbf{\textsc{Overall~($\uparrow$)}}} \\
  
  \midrule
  ACT
  & 8.57\%
  & 16.7$\pm$6.2\%
  & 18.3$\pm$4.7\%
  & 56.7$\pm$6.2\%
  & 30.56$\pm$19.36\%
  & 78.3$\pm$9.4\%
  & 73.3$\pm$8.5\%
  & 56.7$\pm$6.2\%
  & 33.3$\pm$2.4\%
  & 60.42$\pm$18.98\%   & 47.61\% \\

  DP
  & 8.33\%
  & \textbf{45.0$\pm$10.8\%}
  & 36.7$\pm$4.7\%
  & 75.0$\pm$4.1\%
  & \textbf{52.22$\pm$17.97\%}
  & 85.0$\pm$10.8\%
  & 78.3$\pm$14.3\%
  & \textbf{76.7$\pm$2.4\%}
  & 23.3$\pm$6.2\%
  & 65.83$\pm$26.52\%   & 60\% \\

  FM
  & 5.71\%
  & 13.3$\pm$2.4\%
  & 38.3$\pm$4.7\%
  & 73.3$\pm$6.2\%
  & 41.67$\pm$25.06\%
  & 70.0$\pm$4.1\%
  & 15.0$\pm$14.7\%
  & 70.0$\pm$8.2\%
  & 38.3$\pm$14.3\%
  & 48.33$\pm$26.94\%   & 45.46\% \\

  $\pi_{0.5}$
  & \underline{5.24\%}
  & 10.0$\pm$4.1\%
  & \textbf{43.3$\pm$6.2\%}
  & 71.7$\pm$11.8\%
  & 41.67$\pm$26.46\%
  & 66.7$\pm$6.2\%
  & 73.3$\pm$2.4\%
  & 70.0$\pm$0.0\%
  & 23.3$\pm$6.2\%
  & 58.33$\pm$20.85\%   & 51.19\% \\

  \rowcolor{hclbrown!8!white}
  \texttt{Ours}
  & \textbf{4.29\%}
  & \underline{28.3$\pm$6.2\%}
  & \underline{41.7$\pm$2.4\%}
  & \textbf{80.0$\pm$4.1\%}
  & \underline{50.00$\pm$22.36\%}
  & \underline{90.0$\pm$4.1\%}
  & \textbf{100.0$\pm$0.0\%}
  & \underline{73.3$\pm$13.1\%}
  & \underline{75.0$\pm$10.8\%}
  & \textbf{84.58$\pm$14.06\%}   & \textbf{69.76\%} \\

  \rowcolor{hclbrown!15!white}
  \texttt{Ours}$^{*}$
  & \textbf{4.29\%}
  & 26.7$\pm$4.7\%
  & 28.3$\pm$2.4\%
  & \underline{78.3$\pm$8.5\%}
  & 44.44$\pm$24.66\%
  & \textbf{93.3$\pm$2.4\%}
  & \underline{96.7$\pm$2.4\%}
  & 70.0$\pm$8.2\%
  & \textbf{76.7$\pm$18.9\%}
  & \underline{84.17$\pm$15.25\%}   & \underline{67.14\%} \\

  \bottomrule
  \end{tabular}
  \end{adjustbox}
\end{table}

%% file: table/arena_multi_task.tex
\begin{table}[H]
  \caption{
    \textbf{\textit{Multi-task results on HumanoidArena.}}
    Each policy is trained jointly on the seven benchmark tasks.
    ACT, DP, FM, and $\pi_{0.5}$ results are taken from the official HumanoidArena report~\citep{humanoidarena}. HuGe results are from Table~4 of \citet{liu2026huge}. Entries marked $^{\dagger}$ are from Table~1 of \citet{li2026languageforcing}, with HOI and HSI means computed from the reported task scores.
    $^{*}$ indicates additional policy pretraining, with data and procedures specific to each method. We show mean $\pm$ standard deviation where available and use a dash for unreported values. Bold and underlined entries indicate the best and second-best results, respectively.
   Baselines include ACT~\citep{ACT}, DP~\citep{chi2023diffusionpolicy}, FM~\citep{FM_objective}, $\pi_{0.5}$~\citep{pi05}, HuGe variants~\citep{liu2026huge}, Fast-WAM~\citep{yuan2026fastwam}, LingBot-VA~\citep{li2026lingbotva}, GR00T N1.7~\citep{grootn1}, GigaWorld-Policy~\citep{gigaworld2026policy05}.
  }
  \label{tab:arena_multi_task}
  \centering
  \arrayrulecolor{hclblue}
  \begin{adjustbox}{max width=\textwidth}
  \setlength{\tabcolsep}{4.2pt}
  \renewcommand{\arraystretch}{1.08}

  \begin{tabular}{cccccc|cccccc}
  \toprule
  & &
  \multicolumn{4}{c|}{\textbf{Human--Object Interaction}} &
  \multicolumn{5}{c}{\textbf{Human--Scene Interaction}} & \\
  \cmidrule(lr){3-6}
  \cmidrule(lr){7-11}

  \multirow{-2}{*}{\textbf{Method}} &
  \multirow{-2}{*}{\textbf{AFR~($\downarrow$)}} &
  \textbf{\textsc{Football}} &
  \textbf{\textsc{DoubleDesk}} &
  \textbf{\textsc{P\&Pbox}} &
  \textbf{\textsc{Avg~($\uparrow$)}} &
  \textbf{\textsc{OpenDoor}} &
  \textbf{\textsc{SitSofa}} &
  \textbf{\textsc{Boxing}} &
  \textbf{\textsc{VisNavi}} &
  \textbf{\textsc{Avg~($\uparrow$)}} &
  \multirow{-2}{*}{\textbf{\textsc{Overall~($\uparrow$)}}} \\
  \midrule

  ACT
  & 7.4\%
  & 20.0$\pm$4.1\%
  & 30.0$\pm$4.1\%
  & 60.0$\pm$12.2\%
  & 36.7$\pm$18.7\%
  & 58.3$\pm$2.4\%
  & 21.7$\pm$2.4\%
  & 58.3$\pm$10.3\%
  & 28.3$\pm$4.7\%
  & 41.7$\pm$17.8\%
  & 39.5\% \\

  DP
  & 7.1\%
  & 53.3$\pm$6.2\%
  & 35.0$\pm$10.8\%
  & 0.0$\pm$0.0\%
  & 29.4$\pm$23.3\%
  & 83.3$\pm$6.2\%
  & \underline{83.3$\pm$4.7\%}
  & 21.7$\pm$30.6\%
  & 50.0$\pm$14.7\%
  & 59.6$\pm$31.1\%
  & 46.7\% \\

  FM
  & 7.1\%
  & 21.7$\pm$8.5\%
  & 20.0$\pm$7.1\%
  & 0.0$\pm$0.0\%
  & 13.9$\pm$11.7\%
  & 76.7$\pm$2.4\%
  & 48.3$\pm$6.2\%
  & 18.3$\pm$25.9\%
  & 38.3$\pm$6.2\%
  & 45.4$\pm$25.1\%
  & 31.9\% \\

  $\pi_{0.5}$
  & 6.4\%
  & 10.0$\pm$7.1\%
  & 30.0$\pm$7.1\%
  & 75.0$\pm$4.1\%
  & 38.3$\pm$27.9\%
  & 85.0$\pm$4.1\%
  & 70.0$\pm$14.7\%
  & 71.7$\pm$12.5\%
  & 35.0$\pm$4.1\%
  & 65.4$\pm$21.1\%
  & 53.8\% \\

  HuGe-OFT$^{*}$
  & ---
  & 48.33\%
  & 0.00\%
  & 53.33\%
  & 33.89\%
  & 75.00\%
  & 55.00\%
  & \underline{73.33\%}
  & \underline{73.33\%}
  & 69.17\%
  & 54.05\% \\

  HuGe-GR00T$^{*}$
  & ---
  & \textbf{71.67\%}
  & 15.00\%
  & \underline{81.67\%}
  & 56.11\%
  & \textbf{95.00\%}
  & 78.33\%
  & 68.33\%
  & 56.67\%
  & 74.58\%
  & 66.67\% \\

  HuGe-Pi$^{*}$
  & ---
  & 63.33\%
  & 11.67\%
  & 76.67\%
  & 50.56\%
  & 88.33\%
  & \underline{83.33\%}
  & 68.33\%
  & 63.33\%
  & 75.83\%
  & 65.00\% \\

  Fast-WAM$^{\dagger}$
  & ---
  & 46.25\%
  & 45.00\%
  & 73.75\%
  & 55.00\%
  & 85.42\%
  & 74.58\%
  & 67.92\%
  & 35.00\%
  & 65.73\%
  & 61.13\% \\

  LingBot-VA$^{\dagger}$
  & ---
  & 51.67\%
  & 41.25\%
  & 77.92\%
  & 56.95\%
  & 84.58\%
  & 77.08\%
  & 72.92\%
  & 52.08\%
  & 71.67\%
  & 65.36\% \\

  GR00T N1.7$^{\dagger}$
  & ---
  & \underline{63.75\%}
  & \underline{49.58\%}
  & 73.33\%
  & \textbf{62.22\%}
  & 82.08\%
  & 76.67\%
  & \textbf{76.67\%}
  & 65.83\%
  & 75.31\%
  & \underline{69.70\%} \\

  GigaWorld-Policy$^{\dagger}$
  & ---
  & 62.08\%
  & \textbf{52.50\%}
  & 65.83\%
  & \underline{60.14\%}
  & 90.83\%
  & 74.17\%
  & \underline{73.33\%}
  & 60.00\%
  & 74.58\%
  & 68.39\% \\

  \rowcolor{hclbrown!8!white}
  \texttt{Ours}
  & \underline{5.0\%}
  & 35.0$\pm$7.1\%
  & 40.0$\pm$0.0\%
  & 80.0$\pm$4.1\%
  & 51.7$\pm$20.7\%
  & \underline{93.3$\pm$6.2\%}
  & \textbf{96.7$\pm$4.7\%}
  & 66.7$\pm$4.7\%
  & 61.7$\pm$6.2\%
  & \underline{79.6$\pm$16.5\%}
  & 67.6\% \\

  \rowcolor{hclbrown!15!white}
  \texttt{Ours}$^{*}$
  & \textbf{4.3\%}
  & 40.0$\pm$4.1\%
  & 38.3$\pm$2.4\%
  & \textbf{90.0$\pm$4.1\%}
  & 56.1$\pm$24.2\%
  & \textbf{95.0$\pm$4.1\%}
  & \textbf{96.7$\pm$4.7\%}
  & 68.3$\pm$10.3\%
  & \textbf{76.7$\pm$8.5\%}
  & \textbf{84.2$\pm$14.1\%}
  & \textbf{72.1\%} \\

  \bottomrule
  \end{tabular}
  \end{adjustbox}
\end{table}

%% file: table/simple_single_task.tex
\begin{table}[!b]
  \caption{
    \textbf{\textit{Single-Task Results on SIMPLE.}} L0, L1, and L2 denote evaluation settings with progressively stronger visual and lighting variation and target-pose randomization, from the least perturbed setting (L0) to the most perturbed setting (L2). Each entry reports the number of successful trials at L0/L1/L2, followed in parentheses by the aggregate success rate across all three levels. Baseline results in the first three task columns are from the official SIMPLE repository; the last five $\Psi_0$ results are from Table~7 of the initial SIMPLE report~\citep{simple}. For ACT, $\pi_{0.5}$, EgoVLA, InternVLA-M1, and DP, results in the last five task columns are taken from Table~I of \citet{chen2026memorybeforegoal}. Avg is the mean success rate across all eight tasks, computed from the trial counts. Baseline rows compile results from these sources rather than a unified reevaluation; training and execution settings follow the respective reports. Rows with missing task results show a dash. Bold and underlined values mark the best and second-best aggregate success rates, respectively.
   Baselines include ACT~\citep{ACT}, DreamZero~\citep{dreamzero}, $\pi_{0.5}$~\citep{pi05}, EgoVLA~\citep{egovla}, GR00T N1.6~\citep{grootn1}, InternVLA-M1~\citep{internvla_m1}, DP~\citep{chi2023diffusionpolicy}, H-RDT~\citep{hrdt}, $\Psi_0$~\citep{psi0}.
  }
  \label{table:simple_main_results}
  \centering
  \arrayrulecolor{hclblue}
  \setlength{\tabcolsep}{0pt}
  \renewcommand{\arraystretch}{1.08}
  \fontsize{6.2}{7}\selectfont
  \newcommand{\taskheader}[1]{{\fontsize{5.55}{6.2}\selectfont\textbf{\textsc{\hspace{2.2pt}#1\hspace{2.2pt}}}}}
  \newcommand{\resultentry}[2]{{\fontsize{5.2}{6}\selectfont\hspace{2.2pt}#1\;(#2)\hspace{2.2pt}}}
  \newcommand{\bestentry}[2]{{\fontsize{5.2}{6}\selectfont\textbf{\hspace{2.2pt}#1\;(#2)\hspace{2.2pt}}}}
  \newcommand{\secondentry}[2]{{\fontsize{5.2}{6}\selectfont\hspace{2.2pt}\underline{#1\;(#2)}\hspace{2.2pt}}}

  \newcommand{\avgentry}[1]{{\fontsize{5.2}{6}\selectfont\hspace{2.2pt}#1\hspace{2.2pt}}}
  \begin{adjustbox}{width=\textwidth}
  \begin{tabular}{@{}c*{9}{c}@{}}
    \toprule
    {\fontsize{5.55}{6.2}\selectfont\textbf{Method}}
    & \taskheader{Handover}
    & \taskheader{Mobile P\&P}
    & \taskheader{XMoveBendPick}
    & \taskheader{CloseDoor}
    & \taskheader{OpenOven}
    & \taskheader{OpenFaucet}
    & \taskheader{Push Office Chair}
    & \taskheader{OpenTrashCan}
    & \taskheader{Avg ($\uparrow$)} \\
    \midrule

    ACT
    & \resultentry{4/4/6}{47\%}
    & \secondentry{6/5/7}{60\%}
    & \resultentry{6/8/8}{73\%}
    & \bestentry{10/10/10}{100\%} & \resultentry{1/0/1}{7\%} & \resultentry{0/0/1}{3\%} & \secondentry{9/10/10}{97\%} & \resultentry{4/4/7}{50\%}
    & \avgentry{54.58\%} \\

    DreamZero
    & \secondentry{7/8/9}{80\%}
    & \resultentry{5/3/3}{37\%}
    & \resultentry{0/0/1}{3\%}
    & - & - & - & - & -
    & \avgentry{-} \\

    $\pi_{0.5}$
    & \resultentry{5/4/5}{47\%}
    & \resultentry{3/3/3}{30\%}
    & \resultentry{0/0/0}{0\%}
    & \secondentry{10/10/9}{97\%} & \resultentry{4/3/6}{43\%} & \resultentry{4/2/0}{20\%} & \resultentry{10/10/6}{87\%} & \resultentry{7/6/5}{60\%}
    & \avgentry{47.92\%} \\

    EgoVLA
    & \resultentry{0/4/3}{23\%}
    & \resultentry{0/0/0}{0\%}
    & \resultentry{3/5/4}{40\%}
    & \resultentry{10/8/10}{93\%} & \resultentry{5/8/2}{50\%} & \resultentry{0/0/0}{0\%} & \secondentry{10/9/10}{97\%} & \resultentry{3/5/4}{40\%}
    & \avgentry{42.92\%} \\

    GR00T N1.6
    & \resultentry{1/3/3}{23\%}
    & \resultentry{0/0/0}{0\%}
    & \resultentry{4/4/1}{30\%}
    & - & - & - & - & -
    & \avgentry{-} \\

    InternVLA-M1
    & \resultentry{0/0/0}{0\%}
    & \resultentry{0/0/0}{0\%}
    & \resultentry{3/5/7}{50\%}
    & \resultentry{0/0/0}{0\%} & \bestentry{8/6/8}{73\%} & \resultentry{0/0/0}{0\%} & \resultentry{2/3/1}{20\%} & \resultentry{0/0/0}{0\%}
    & \avgentry{17.92\%} \\

    DP
    & \resultentry{3/2/4}{30\%}
    & \resultentry{4/0/0}{13\%}
    & \resultentry{0/0/0}{0\%}
    & \bestentry{10/10/10}{100\%} & \resultentry{0/0/0}{0\%} & \secondentry{2/3/6}{37\%} & \resultentry{10/6/8}{80\%} & \resultentry{4/7/7}{60\%}
    & \avgentry{40.00\%} \\

    H-RDT
    & \resultentry{0/1/0}{3\%}
    & \resultentry{0/0/0}{0\%}
    & \resultentry{0/0/0}{0\%}
    & - & - & - & - & -
    & \avgentry{-} \\

    $\Psi_0$
    & \secondentry{7/7/10}{80\%}
    & \secondentry{7/5/6}{60\%}
    & \bestentry{10/9/9}{93\%}
    & \bestentry{10/10/10}{100\%}
    & \resultentry{7/5/4}{53\%}
    & \resultentry{3/3/4}{33\%}
    & \bestentry{10/10/10}{100\%}
    & \bestentry{7/9/9}{83\%}
    & \avgentry{\underline{75.42\%}} \\

    \rowcolor{hclbrown!15!white}
    \texttt{Ours}$^{*}$
    & \bestentry{9/8/8}{83\%}
    & \bestentry{9/8/4}{70.0\%}
    & \secondentry{10/7/7}{80\%}
    & \bestentry{10/10/10}{100\%}
    & \secondentry{8/4/5}{57\%}
    & \bestentry{5/6/5}{53\%}
    & \bestentry{10/10/10}{100\%}
    & \secondentry{9/4/6}{63\%}
    & \avgentry{\textbf{75.83\%}} \\

    \bottomrule
  \end{tabular}
  \end{adjustbox}
\end{table}

%% file: table/proof_t2m.tex
\begin{table}[H]
  \caption{\textbf{Effect of Motion-Prior Selection at Matched Backbone Capacity.} Two 282M-parameter Kimodo-G1~\citep{kimodo} backbones are adapted using the same policy architecture and downstream protocol. Data durations refer to the full motion corpora of the released checkpoints, including their training and test splits. Task scores are reported as mean $\pm$ standard deviation over three seeds. The comparison varies the pretrained motion corpus, including both its size and content.}
  \label{tab:t2m_dependency}
  \centering
  \arrayrulecolor{hclblue}
  \begin{adjustbox}{max width=\textwidth}
  \setlength{\tabcolsep}{3.8pt}
  \renewcommand{\arraystretch}{1.08}

  \begin{tabular}{cccccc|ccccc}
  \toprule
  \multirow{2}{*}{\textbf{Base Model}} &
  \multirow{2}{*}{\textbf{Model Size}} &
  \multirow{2}{*}{\textbf{Data Scale}} &
  \multicolumn{3}{c|}{\textbf{Human--Object Interaction}} &
  \multicolumn{4}{c}{\textbf{Human--Scene Interaction}} &
  \multirow{2}{*}{\textbf{Overall~($\uparrow$)}} \\
  \cmidrule(lr){4-6}
  \cmidrule(lr){7-10}

  & & &
  \textbf{\textsc{Football}} &
  \textbf{\textsc{DoubleDesk}} &
  \textbf{\textsc{P\&Pbox}} &
  \textbf{\textsc{OpenDoor}} &
  \textbf{\textsc{SitSofa}} &
  \textbf{\textsc{Boxing}} &
  \textbf{\textsc{VisNavi}} \\
  \midrule

  \texttt{Kimodo-G1-SEED-v1}
  & 282\,M
  & 288\,h
  & \textbf{50.00$\pm$16.33\%}
  & \underline{33.33$\pm$15.46\%}
  & \underline{75.00$\pm$10.80\%}
  & \underline{60.00$\pm$4.08\%}
  & \underline{90.00$\pm$4.08\%}
  & \textbf{71.67$\pm$8.50\%}
  & \textbf{70.00$\pm$18.71\%}
  & \underline{64.29\%} \\

  \rowcolor{hclbrown!8!white}
  \texttt{Kimodo-G1-RP-v1}
  & 282\,M
  & $\sim$560\,h
  & \underline{35.0$\pm$7.1\%}
  & \textbf{40.0$\pm$0.0\%}
  & \textbf{80.0$\pm$4.1\%}
  & \textbf{93.3$\pm$6.2\%}
  & \textbf{96.7$\pm$4.7\%}
  & \underline{66.7$\pm$4.7\%}
  & \underline{61.7$\pm$6.2\%}
  & \textbf{67.63\%} \\

  \bottomrule
  \end{tabular}
  \end{adjustbox}
\end{table}

%% file: content/Conclusion.tex
\section{Conclusion}
\label{sec:conclusion}

We presented Kinematic Action Models (KAMs), which use a pretrained motion generator as the primary backbone for action generation, and Kimodo-Policy as an initial instantiation for humanoid loco-manipulation. Kimodo-Policy retains a frozen motion backbone and learns visual conditioning and hand control around its root--body generation hierarchy. Motion-history constraints and repeated action-chunk prediction ground this inherited motion structure in ongoing execution, connecting motion synthesis to closed-loop interaction. Experiments on HumanoidArena, SIMPLE, and eight real-world tasks support the feasibility of this approach with 92M trainable parameters out of 678M total parameters. Ablations support retaining the pretrained motion backbone, while the motion-prior comparison shows that downstream performance depends on the selected prior at matched backbone capacity. Additional robot-interaction pretraining benefits joint multi-task training and real-world deployment, although it does not improve the aggregate single-task HumanoidArena result. Data and capacity sweeps further show gains from strengthening adaptation around a fixed backbone within the evaluated range. These findings provide initial evidence for KAM-based humanoid control.

\textbf{Limitations.} A principal limitation of the current framework is its dependence on a specific low-level motion tracker. Kimodo-Policy is designed and evaluated with SONIC, and its motion-reference interface and deployment pipeline are tailored to this tracker; other tracking controllers are not currently supported. Consequently, the reported results establish the effectiveness of the policy within the SONIC-based control stack, while transfer across trackers remains unverified. Although separating high-level motion generation from low-level execution provides a modular organization, this separation alone does not ensure compatibility with controllers that differ in reference parameterization, tracking behavior, or executable motion range. Extending the framework would therefore require adapting the policy–tracker interface and validating closed-loop performance under the target controller. Developing a more general interface between pretrained motion generation and low-level whole-body control remains an important direction for broadening its applicability. The present evidence concerns one KAM instantiation; generalization across motion generators and broader interaction settings remains to be established.

\textbf{Future Work.} A natural extension is to combine the motion-generation backbone with richer semantic and predictive representations. Vision--language and video models could condition the KAM on task-level context and anticipated environmental changes while retaining motion generation as the primary action mechanism. A complementary direction is to improve the motion prior through broader motion coverage and more capable generators. The current comparison motivates studying this direction, but does not separate the effects of motion-data size, data composition, and generation quality. Human motion resources offer an additional route to pretraining, through retargeted robot-space trajectories or transfer from pretrained human motion generators. Such transfer requires aligning representations across embodiments and accounting for robot-specific actuation, balance, and contact constraints. Together with a more general policy--tracker interface, these extensions could broaden the range of closed-loop behaviors supported by KAMs.